# A New Technique for INS/GNSS Attitude and Parameter Estimation Using Online Optimization

Yuanxin Wu, Jinling Wang and Dewen Hu

**Abstract—Integration of inertial navigation system (INS) and global navigation satellite system (GNSS) is usually implemented in engineering applications by way of Kalman-like filtering. This form of INS/GNSS integration is prone to attitude initialization failure, especially when the host vehicle is moving freely. This paper proposes an online constrained-optimization method to simultaneously estimate the attitude and other related parameters including GNSS antenna's lever arm and inertial sensor biases. This new technique benefits from self-initialization in which no prior attitude or sensor measurement noise information is required. Numerical results are reported to validate its effectiveness and prospect in high accurate INS/GNSS applications.**

## I. Introduction

Inertial navigation system (INS) is versatile in providing rich kinds of motion information of the host vehicle (called 'carrier' hereafter). Global navigation satellite system (GNSS) is widely used outdoors as a convenient velocity and position sensor. INS/GNSS integration has become the most common means for both civilian and military navigation applications, because of their combined benefits of continuous output, guaranteed accuracy and low opportunity of interference or jamming.

Reference frame unification is a pre-condition for any accurate information blending of multiple sensors.

This work was supported in part by the Fok Ying Tung Foundation (131061), National Natural Science Foundation of China (61174002), the Foundation for the Author of National Excellent Doctoral Dissertation of People's Republic of China (FANEDD 200897) and Program for New Century Excellent Talents in University (NCET-10-0900).

Authors' address: Y. Wu, School of Aeronautics and Astronautics, Central South University, Changsha, Hunan, China, 410083. Tel: 086-0731-88877132, E-mail: (yuanx_wu@hotmail.com); J. Wang, School of Civil and Environmental Engineering, University of New South Wales, E-mail: (jinling.wang@unsw.edu.au); D. Hu, Department of Automatic Control, College of Mechatronics and Automation, National University of Defense Technology, Changsha, Hunan, China, 410073, E-mail: (dwhu@nudt.edu.cn).

For the case of INS/GNSS integration, there is an unavoidable displacement, usually called as the lever arm, between their respective sensing points of INS and GNSS [1]. The magnitude of the lever arm is significant for many applications. For example, when the GNSS antenna is mounted outside on the roof to better receive satellite signals, the INS is usually installed inside the carrier for reasons like easy maintenance. It is difficult or impossible, however, to precisely measure the lever arm vector in a designated frame, taking account of the virtual attributes of the sensing points and the frame axes [2, 3]. Lever arm uncertainty acts as a major error source of INS/GNSS in accurate applications, e.g., pose determination in airborne direct georeferencing [4] and airborne gravity measuring [5].

A good practice is to accommodate and estimate the lever arm within the extended Kalman filter (EKF) in INS/GNSS integration [2, 6, 7]. In addition to the knowledge of noise statistics, the Kalman filtering method relies heavily on a roughly known initial attitude that is only achievable under benign situations such as when the carrier is stationary or moving straight. If not properly initiated, however, INS/GNSS integration based on Kalman filtering would be subjected to failure [1, 8, 9]. These situations are not uncommon in practice, for example when the carrier is moving freely or the duration of benign situations is not long enough to reach a good initial attitude alignment. That is to say, the lever arm estimation is conditioned on a good initial attitude. On the other hand, to obtain a good initial attitude inversely requires a small INS/GNSS lever arm effect. Our recent work of in-flight coarse alignment [10, 11] clearly shows that the presence of the lever arm causes slow convergence and degradation of attitude estimation. It is desirable to find a way to simultaneously solve the problems of the lever arm and attitude alignment, for a wide spectrum of INS/GNSS application scenarios.

In contrast to the Kalman filtering, a novel optimization-based approach to solve attitude alignment was initiated by our group [12] in the swaying case and recently extended to the on-the-move alignment by us and other groups [10, 13-15]. The approach poses the attitude alignment as an attitude optimization problem using infinite vector observations and is advantageous in handling those scenarios that have no prior initial attitude information available. It has the benefit of an exact eigenvalue/eigenvector solution, but in principle is not capable of estimating parameters other than attitude, such as the sensor bias and the GNSS lever arm mentioned above [10, 12].

This paper further traces the thread of optimization-based approach by addressing joint estimation of attitude and the associated parameters. The contents are organized as follows. Section II formulates the joint estimation problem as a nonlinear constrained optimization on attitude, gyroscope/accelerometer biases and the lever arm. Section III presents approximate algorithms to compute the integrals involved in the problem. Section IV uses the iterative Newton-Lagrange method to solve the nonlinear constrained optimization. A time-recursive algorithm is derived by employing the special property of the problem. Section V reports simulation results of the algorithm and compares with a properly-initiated and well-tuned EKF. Conclusions are finally drawn in Section VI. The main contribution of this paper is a novel online processing technique for INS/GNSS integration, derived from nonlinear constrained optimization, which can estimate other parameters other than attitude. In contrast to those Kalman-like filtering methods, the new technique does not require any prior initialization information to work, for example, the initial attitude value or sensor noise characteristics.

## II. Problem Formulation

Denote by $N$ the local level reference frame, by $B$ the INS body frame, by $I$ the inertial non-rotating frame, and by $E$ the Earth frame. The navigation (attitude, velocity and position) rate equations in the reference $N$-frame are well known as [1, 16, 17]

$$\dot{\mathbf{C}}_b^n = \mathbf{C}_b^n \left( \boldsymbol{\omega}_{nb}^b \times \right), \quad \boldsymbol{\omega}_{nb}^b = \boldsymbol{\omega}_{ib}^b - \mathbf{b}_g - \mathbf{C}_n^b \left( \boldsymbol{\omega}_{ie}^n + \boldsymbol{\omega}_{en}^n \right), \tag{1}$$

$$\dot{\mathbf{v}}^n = \mathbf{C}_b^n \left( \mathbf{f}^b - \mathbf{b}_a \right) - \left( 2\boldsymbol{\omega}_{ie}^n + \boldsymbol{\omega}_{en}^n \right) \times \mathbf{v}^n + \mathbf{g}^n \tag{2}$$

$$\dot{\mathbf{p}} = \mathbf{R}_c \mathbf{v}^n \tag{3}$$

where $\mathbf{C}_b^n$ denotes the attitude matrix from the body frame to the reference frame with $\mathbf{C}_n^b = \left( \mathbf{C}_b^n \right)^T$, $\mathbf{v}^n$ the velocity relative to the Earth, $\boldsymbol{\omega}_{ib}^b$ the error-contaminated body angular rate measured by gyroscopes in the body frame, $\mathbf{f}^b$ the error-contaminated specific force measured by accelerometers in the body frame, $\boldsymbol{\omega}_{ie}^n$ the Earth rotation rate with respect to the inertial frame, $\boldsymbol{\omega}_{en}^n$ the angular rate of the reference frame with respect to the Earth frame, $\boldsymbol{\omega}_{nb}^b$ the body angular rate with respect to the reference frame, and $\mathbf{g}^n$ the gravity vector. The $3 \times 3$ skew symmetric matrix $\left( \cdot \times \right)$ is defined so that the cross product satisfies $\mathbf{a}\times\mathbf{b}=(\mathbf{a}\times)\mathbf{b}$ for arbitrary two vectors. The gyroscope bias $\mathbf{b}_g$ and the accelerometer bias $\mathbf{b}_a$ are taken into considerations approximately as random constant.

The position $\mathbf{p}\triangleq[\lambda \quad L \quad h]^T$ is described by the angular orientation of the reference frame relative to the Earth frame, commonly expressed as longitude $\lambda$, latitude $L$ and height $h$ above the Earth surface. The local curvature matrix $\mathbf{R}_c$ is a function of current position. In the context of a specific local level frame choice, e.g., North-Up-East, $\mathbf{v}^n=[v_N \quad v_U \quad v_E]^T$, it is explicitly expressed as

$$\mathbf{R}_c=\begin{bmatrix} 0 & 0 & \frac{1}{(R_E+h)\cos L} \\ \frac{1}{R_N+h} & 0 & 0 \\ 0 & 1 & 0 \end{bmatrix} \tag{4}$$

where $R_E$ and $R_N$ are respectively the transverse radius of curvature and the meridian radius of curvature of the reference ellipsoid. The specific expression of $\mathbf{R}_c$ will be different for other local level frame choices but it does not hinder from understanding the main idea of this paper.

All the quantities herein are functions of time and, if not stated, their time dependences are omitted for brevity.

Suppose the GNSS antenna is rigidly fixed relative to the INS and the lever arm from the INS to the GNSS antenna is expressed in the body frame as $\mathbf{l}^b$. The GNSS antenna's velocity and position are related to the INS position and velocity by [1]

$$\mathbf{p}_{gnss}\approx\mathbf{p}+\mathbf{R}_c\mathbf{C}_b^n\mathbf{l}^b \tag{5}$$

$$\mathbf{v}_{gnss}^n=\mathbf{v}^n+\mathbf{C}_b^n\left(\boldsymbol{\omega}_{eb}^b\times\mathbf{l}^b\right) \tag{6}$$

Next we will try to solve the lever arm using GNSS velocity and position information. Substituting (6) into (2) and using (1)

$$\dot{\mathbf{v}}_{gnss}^n-\mathbf{C}_b^n\boldsymbol{\omega}_{nb}^b\times\left(\boldsymbol{\omega}_{eb}^b\times\mathbf{l}^b\right)-\mathbf{C}_b^n\left(\dot{\boldsymbol{\omega}}_{eb}^b\times\mathbf{l}^b\right)=\mathbf{C}_b^n\left(\mathbf{f}^b-\mathbf{b}_a\right)-\left(2\boldsymbol{\omega}_{ie}^n+\boldsymbol{\omega}_{en}^n\right)\times\left(\mathbf{v}_{gnss}^n-\mathbf{C}_b^n\left(\boldsymbol{\omega}_{eb}^b\times\mathbf{l}^b\right)\right)+\mathbf{g}^n \tag{7}$$

because $\mathbf{l}^b$ is constant. Organizing the terms and using (1)

$$\begin{aligned}\dot{\mathbf{v}}_{gnss}^n+\left(2\boldsymbol{\omega}_{ie}^n+\boldsymbol{\omega}_{en}^n\right)\times\mathbf{v}_{gnss}^n-\mathbf{g}^n&=\mathbf{C}_b^n\left[\mathbf{f}^b-\mathbf{b}_a+\boldsymbol{\omega}_{nb}^b\times\left(\boldsymbol{\omega}_{eb}^b\times\mathbf{l}^b\right)+\dot{\boldsymbol{\omega}}_{eb}^b\times\mathbf{l}^b+\left(2\boldsymbol{\omega}_{ie}^b+\boldsymbol{\omega}_{en}^b\right)\times\left(\boldsymbol{\omega}_{eb}^b\times\mathbf{l}^b\right)\right]\\&=\mathbf{C}_b^n\left[\mathbf{f}^b-\mathbf{b}_a+\dot{\boldsymbol{\omega}}_{eb}^b\times\mathbf{l}^b+\left(\boldsymbol{\omega}_{ie}^b+\boldsymbol{\omega}_{ib}^b-\mathbf{b}_g\right)\times\left(\boldsymbol{\omega}_{eb}^b\times\mathbf{l}^b\right)\right]\end{aligned}\tag{8}$$

We see that the lever arm $\mathbf{l}^b$ is multiplicatively coupled with the attitude matrix $\mathbf{C}_b^n$.

As shown in (1), the attitude matrix $\mathbf{C}_b^n$ is a function of $\boldsymbol{\omega}_{nb}^b$, whose calculation inversely depends on $\mathbf{C}_b^n$. In order to obtain the analytic form of $\mathbf{C}_b^n$, we separately consider the attitude changes of the body frame and the reference frame, both relative to the same chosen inertial frame, and then combine them together [12]. Specifically, by the chain rule of the attitude matrix, $\mathbf{C}_b^n$ at any time satisfies

$$\mathbf{C}_b^n(t)=\mathbf{C}_{b(t)}^{n(t)}=\mathbf{C}_{n(0)}^{n(t)}\mathbf{C}_{b(0)}^{n(0)}\mathbf{C}_{b(t)}^{b(0)}=\mathbf{C}_{n(0)}^{n(t)}\mathbf{C}_b^n(0)\mathbf{C}_{b(t)}^{b(0)}\tag{9}$$

where the initial attitude matrix $\mathbf{C}_b^n(0)$ is constant, and $\mathbf{C}_{b(0)}^{b(t)}$ and $\mathbf{C}_{n(0)}^{n(t)}$, respectively, encode the attitude changes of the body frame and the reference frame from time $0$ to $t$. Their rate equations are

$$\begin{aligned}\dot{\mathbf{C}}_{b(t)}^{b(0)}&=\mathbf{C}_{b(t)}^{b(0)}\left(\boldsymbol{\omega}_{ib}^b-\mathbf{b}_g\right)\times\\\dot{\mathbf{C}}_{n(t)}^{n(0)}&=\mathbf{C}_{n(t)}^{n(0)}\boldsymbol{\omega}_{in}^n\times\end{aligned}\tag{10}$$

where the angular velocity of the reference frame with respect to the inertial frame $\boldsymbol{\omega}_{in}^n=\boldsymbol{\omega}_{ie}^n+\boldsymbol{\omega}_{en}^n$.

Substituting (9) into (8), we have

$$\mathbf{C}_{n(t)}^{n(0)}\left[\dot{\mathbf{v}}_{gnss}^n+\left(2\boldsymbol{\omega}_{ie}^n+\boldsymbol{\omega}_{en}^n\right)\times\mathbf{v}_{gnss}^n-\mathbf{g}^n\right]=\mathbf{C}_b^n(0)\mathbf{C}_{b(t)}^{b(0)}\left[\mathbf{f}^b-\mathbf{b}_a+\dot{\boldsymbol{\omega}}_{eb}^b\times\mathbf{l}^b+\left(\boldsymbol{\omega}_{ie}^b+\boldsymbol{\omega}_{ib}^b-\mathbf{b}_g\right)\times\left(\boldsymbol{\omega}_{eb}^b\times\mathbf{l}^b\right)\right]\tag{11}$$

Integrating (11) on both sides over the time interval of interest $[0,\ t]$,

$$\begin{aligned}&\int_0^t\mathbf{C}_{n(t)}^{n(0)}\dot{\mathbf{v}}_{gnss}^n dt+\int_0^t\mathbf{C}_{n(t)}^{n(0)}\left(2\boldsymbol{\omega}_{ie}^n+\boldsymbol{\omega}_{en}^n\right)\times\mathbf{v}_{gnss}^n dt-\int_0^t\mathbf{C}_{n(t)}^{n(0)}\mathbf{g}^n dt\\&=\mathbf{C}_b^n(0)\left[\int_0^t\mathbf{C}_{b(t)}^{b(0)}\left(\mathbf{f}^b-\mathbf{b}_a\right)dt+\int_0^t\mathbf{C}_{b(t)}^{b(0)}\left(\dot{\boldsymbol{\omega}}_{eb}^b\times+\left(\left(\boldsymbol{\omega}_{ie}^b+\boldsymbol{\omega}_{ib}^b-\mathbf{b}_g\right)\times\right)\left(\boldsymbol{\omega}_{eb}^b\times\right)\right)dt\,\mathbf{l}^b\right]\end{aligned}\tag{12}$$

The integral on the left $\int_0^t\mathbf{C}_{n(t)}^{n(0)}\dot{\mathbf{v}}_{gnss}^n dt$ is developed as

$$\int_0^t\mathbf{C}_{n(t)}^{n(0)}\dot{\mathbf{v}}_{gnss}^n dt=\mathbf{C}_{n(t)}^{n(0)}\mathbf{v}_{gnss}^n\Big|_0^t-\int_0^t\mathbf{C}_{n(t)}^{n(0)}\boldsymbol{\omega}_{in}^n\times\mathbf{v}_{gnss}^n dt=\mathbf{C}_{n(t)}^{n(0)}\mathbf{v}_{gnss}^n-\mathbf{v}_{gnss}^n(0)-\int_0^t\mathbf{C}_{n(t)}^{n(0)}\boldsymbol{\omega}_{in}^n\times\mathbf{v}_{gnss}^n dt\tag{13}$$

where the attitude rate equation (10) is used and $\mathbf{v}_{gnss}^{n}(0)$ is the initial GNSS velocity. The integral on the right $\int_0^t \mathbf{C}_{b(t)}^{b(0)} \dot{\boldsymbol{\omega}}_{eb}^{b} \times dt$ is developed as

$$\int_0^t \mathbf{C}_{b(t)}^{b(0)} \dot{\boldsymbol{\omega}}_{eb}^{b} \times dt = \mathbf{C}_{b(t)}^{b(0)} \boldsymbol{\omega}_{eb}^{b} \times - \boldsymbol{\omega}_{eb}^{b}(0) \times - \int_0^t \mathbf{C}_{b(t)}^{b(0)} \left( \left( \boldsymbol{\omega}_{ib}^{b} - \mathbf{b}_g \right) \times \right) \left( \boldsymbol{\omega}_{eb}^{b} \times \right) dt \tag{14}$$

Substituting (13)-(14) into (12), we obtain

$$\begin{aligned}
\boldsymbol{\beta} &\triangleq \mathbf{C}_{n(t)}^{n(0)} \mathbf{v}_{gnss}^{n} - \mathbf{v}_{gnss}^{n}(0) + \int_0^t \mathbf{C}_{n(t)}^{n(0)} \boldsymbol{\omega}_{ie}^{n} \times \mathbf{v}_{gnss}^{n} dt - \int_0^t \mathbf{C}_{n(t)}^{n(0)} \mathbf{g}^{n} dt \\
&= \mathbf{C}_b^n(0) \left[ \int_0^t \mathbf{C}_{b(t)}^{b(0)} \left( \mathbf{f}^b - \mathbf{b}_a \right) dt + \left( \mathbf{C}_{b(t)}^{b(0)} \boldsymbol{\omega}_{eb}^{b} \times - \boldsymbol{\omega}_{eb}^{b}(0) \times + \int_0^t \mathbf{C}_{b(t)}^{b(0)} \left( \boldsymbol{\omega}_{ie}^{b} \times \right) \left( \boldsymbol{\omega}_{eb}^{b} \times \right) dt \right) \mathbf{l}^b \right] \\
&= \mathbf{C}_b^n(0) \left[ \int_0^t \mathbf{C}_{b(t)}^{b(0)} \left( \mathbf{f}^b - \mathbf{b}_a \right) dt + \left( \mathbf{C}_{b(t)}^{b(0)} \boldsymbol{\omega}_{eb}^{b} \times - \boldsymbol{\omega}_{eb}^{b}(0) \times + \left( \boldsymbol{\omega}_{ie}^{b(0)} \times \right) \int_0^t \mathbf{C}_{b(t)}^{b(0)} \left( \boldsymbol{\omega}_{eb}^{b} \times \right) dt \right) \mathbf{l}^b \right] \\
&\approx \mathbf{C}_b^n(0) \left[ \int_0^t \mathbf{C}_{b(t)}^{b(0)} \left( \mathbf{f}^b - \mathbf{b}_a \right) dt + \left( \mathbf{C}_{b(t)}^{b(0)} \boldsymbol{\omega}_{ib}^{b} \times - \boldsymbol{\omega}_{ib}^{b}(0) \times \right) \mathbf{l}^b \right]
\end{aligned} \tag{15}$$

where the newly defined time-varying vector $\boldsymbol{\beta}$ is a function of GNSS position and velocity, while the right side of (15) is a function of error-contaminated gyroscope/accelerometer output. The approximation above is reasonable, since the Earth angular rate $\boldsymbol{\omega}_{ie}^{b}$ is negligible in practice with respect to the INS body-related angular rates $\boldsymbol{\omega}_{ib}^{b}$ or $\boldsymbol{\omega}_{eb}^{b}$.

*Remark 2.1*: It will be a good approximation if the condition is satisfied, i.e., $\left\| \mathbf{C}_{b(t)}^{b(0)} \boldsymbol{\omega}_{eb}^{b} \times \right\| \gg \left\| \left( \boldsymbol{\omega}_{ie}^{b(0)} \times \right) \int_0^t \mathbf{C}_{b(t)}^{b(0)} \left( \boldsymbol{\omega}_{eb}^{b} \times \right) dt \right\|$ in magnitude. As the Earth rotation rate is small ($\sim 7.3\times10^{-5}$ rad/s), the time duration for it being a valid approximation is no less than thousands of seconds.

*Remark 2.2*: The integral term $\int_0^t \mathbf{C}_{b(t)}^{b(0)} \left( \mathbf{f}^b - \mathbf{b}_a \right) dt$ is generally increasing in magnitude as time goes, in contrast to the coefficient term of the lever arm, so it can be inferred that the lever arm effect would be significantly mitigated after a while. It accords with our previous observation in [10] where the lever arm was not considered at all. This property offers us a good rough attitude estimate to be used as an initial value for the optimization problem in Section IV.

The integration equation (15) is explicitly a system of functions of such unknowns as the initial attitude matrix $\mathbf{C}_b^n(0)$, the lever arm $\mathbf{l}^b$ and the accelerometer bias $\mathbf{b}_a$. Regarding (10), it is also an implicit equation of the unknown gyroscope bias $\mathbf{b}_g$ due to the presence of $\mathbf{C}_{b(t)}^{b(0)}$. The noncommutativity property of finite rotations [18] determines that attitude has no explicit form in terms of the gyroscope bias, which makes the gyroscope bias the most difficult to estimate among the four parameters. Interesting to note that (15) is in form quite similar to the famous hand-eye calibration problem in robotics [19].

## III. Approximate Integral Computation

We calculate the integrals in (15) using the similar technique as in [10] and then solve the system of equations (15) in next section. Suppose the current time $t$ is an integer, say *M*, times of the updated interval, i.e., $t \triangleq MT$, where $T$ is the uniform time duration of the update interval $[t_k \quad t_{k+1}]$ with $t_k = kT$.

The second integral of $\boldsymbol{\beta}$ in (15) can be written as

$$\int_0^t \mathbf{C}_{n(t)}^{n(0)} \mathbf{g}^n dt = \sum_{k=0}^{M-1} \int_{t_k}^{t_{k+1}} \mathbf{C}_{n(t)}^{n(0)} \mathbf{g}^n dt = \sum_{k=0}^{M-1} \mathbf{C}_{n(t_k)}^{n(0)} \int_{t_k}^{t_{k+1}} \mathbf{C}_{n(t)}^{n(t_k)} \mathbf{g}^n dt \tag{16}$$

Because $\boldsymbol{\omega}_{in}^n$ is usually slowly changing, we approximate the attitude matrix by $\mathbf{C}_{n(t)}^{n(t_k)} = I + \frac{\sin(\|\boldsymbol{\varphi}_n\|)}{\|\boldsymbol{\varphi}_n\|} \boldsymbol{\varphi}_n \times + \frac{1-\cos(\|\boldsymbol{\varphi}_n\|)}{\|\boldsymbol{\varphi}_n\|^2} (\boldsymbol{\varphi}_n \times)^2 \approx \mathbf{I} + \boldsymbol{\varphi}_n \times$, where $\boldsymbol{\varphi}_n \approx \int_{t_k}^t \boldsymbol{\omega}_{in}^n d\tau \approx (t - t_k) \boldsymbol{\omega}_{in}^n$ denotes the *N*-frame rotation vector from $t_k$ to the current time. Then the integral is approximated by

$$\int_0^t \mathbf{C}_{n(t)}^{n(0)} \mathbf{g}^n dt \approx \sum_{k=0}^{M-1} \mathbf{C}_{n(t_k)}^{n(0)} \int_{t_k}^{t_{k+1}} \left( \mathbf{I} + (t - t_k) \boldsymbol{\omega}_{in}^n \times \right) \mathbf{g}^n dt = \sum_{k=0}^{M-1} \mathbf{C}_{n(t_k)}^{n(0)} \left( T\mathbf{I} + \frac{T^2}{2} \boldsymbol{\omega}_{in}^n \times \right) \mathbf{g}^n \tag{17}$$

where the quantities $\boldsymbol{\omega}_{in}^n$ and $\mathbf{g}^n$ can be approximately regarded as constants during the incremental interval and computed using GNSS velocity and position.

Suppose the velocity $\mathbf{v}_{gnss}^n$ changes linearly during $[t_k, \quad t_{k+1}]$,

$$\mathbf{v}_{gnss}^n(t) = \mathbf{v}_{gnss}^n(t_k) + \frac{t - t_k}{T} \left( \mathbf{v}_{gnss}^n(t_{k+1}) - \mathbf{v}_{gnss}^n(t_k) \right) \tag{18}$$

The first integral of $\boldsymbol{\beta}$ is approximated by

$$\begin{aligned}
\int_0^t \mathbf{C}_{n(t)}^{n(0)} \boldsymbol{\omega}_{ie}^n \times \mathbf{v}_{gnss}^n dt &= \sum_{k=0}^{M-1} \mathbf{C}_{n(t_k)}^{n(0)} \int_{t_k}^{t_{k+1}} \mathbf{C}_{n(t)}^{n(t_k)} \boldsymbol{\omega}_{ie}^n \times \mathbf{v}_{gnss}^n dt \\
&\approx \sum_{k=0}^{M-1} \mathbf{C}_{n(t_k)}^{n(0)} \int_{t_k}^{t_{k+1}} \left(\mathbf{I} + (t-t_k)\boldsymbol{\omega}_{in}^n \times\right) \boldsymbol{\omega}_{ie}^n \times \left( \mathbf{v}_{gnss}^n(t_k) + \frac{t-t_k}{T}\left(\mathbf{v}_{gnss}^n(t_{k+1}) - \mathbf{v}_{gnss}^n(t_k)\right)\right) dt \\
&= \sum_{k=0}^{M-1} \mathbf{C}_{n(t_k)}^{n(0)} \left[ \left( T\mathbf{I} + \frac{T^2}{2}\boldsymbol{\omega}_{in}^n \times \right) \boldsymbol{\omega}_{ie}^n \times \mathbf{v}_{gnss}^n(t_k) + \left( \frac{T}{2}\mathbf{I} + \frac{T^2}{3}\boldsymbol{\omega}_{in}^n \times \right) \boldsymbol{\omega}_{ie}^n \times \left(\mathbf{v}_{gnss}^n(t_{k+1}) - \mathbf{v}_{gnss}^n(t_k)\right) \right] \\
&= \sum_{k=0}^{M-1} \mathbf{C}_{n(t_k)}^{n(0)} \left[ \left( \frac{T}{2}\mathbf{I} + \frac{T^2}{6}\boldsymbol{\omega}_{in}^n \times \right) \boldsymbol{\omega}_{ie}^n \times \mathbf{v}_{gnss}^n(t_k) + \left( \frac{T}{2}\mathbf{I} + \frac{T^2}{3}\boldsymbol{\omega}_{in}^n \times \right) \boldsymbol{\omega}_{ie}^n \times \mathbf{v}_{gnss}^n(t_{k+1}) \right]
\end{aligned} \tag{19}$$

where the quantities $\boldsymbol{\omega}_{ie}^n$ is also regarded as constants during the interval of interest.

Substituting (17) and (19),

$$\begin{aligned}
\boldsymbol{\beta}_M &= \mathbf{C}_{n(t_M)}^{n(0)} \mathbf{v}_{gnss}^n - \mathbf{v}_{gnss}^n(0) \\
&+ \sum_{k=0}^{M-1} \mathbf{C}_{n(t_k)}^{n(0)} \left[ \left( \frac{T}{2}\mathbf{I} + \frac{T^2}{6}\boldsymbol{\omega}_{in}^n \times \right) \boldsymbol{\omega}_{ie}^n \times \mathbf{v}_{gnss}^n(t_k) + \left( \frac{T}{2}\mathbf{I} + \frac{T^2}{3}\boldsymbol{\omega}_{in}^n \times \right) \boldsymbol{\omega}_{ie}^n \times \mathbf{v}_{gnss}^n(t_{k+1}) - \left( T\mathbf{I} + \frac{T^2}{2}\boldsymbol{\omega}_{in}^n \times \right) \mathbf{g}^n \right]
\end{aligned} \tag{20}$$

where the subscript $M$ indicates the time dependence of $\boldsymbol{\beta}$ on $t_M$.

Now we turn to the terms on the right side of (15). The matrix $\mathbf{C}_{b(t)}^{b(0)}$ can be approximated to the first order as

$$\begin{aligned}
\mathbf{C}_{b(t)}^{b(0)} &= \mathbf{C}_{b(T)}^{b(0)} \cdots \mathbf{C}_{b(MT)}^{b((M-1)T)} \approx \prod_{i=1}^{M} \left( \mathbf{I} + \boldsymbol{\Theta}_i - T\mathbf{b}_g \times \right) \\
&\approx \tilde{\mathbf{C}}_{b(t)}^{b(0)} - T\sum_{i=1}^{M} \left[ \prod_{j=1}^{i-1} \left( \mathbf{I} + \boldsymbol{\Theta}_j \right) \right] \left( \mathbf{b}_g \times \right) \left[ \prod_{j=i+1}^{M} \left( \mathbf{I} + \boldsymbol{\Theta}_j \right) \right] \\
&\approx \tilde{\mathbf{C}}_{b(t)}^{b(0)} - MT\mathbf{b}_g \times
\end{aligned} \tag{21}$$

where $\boldsymbol{\Theta}_i$ is the skew symmetric matrix formed by the erroneous incremental rotation vector during the update interval $\left[t_{i-1} \quad t_i\right]$ and $\tilde{\mathbf{C}}_{b(t)}^{b(0)}$ denotes the erroneous body matrix computed by $\boldsymbol{\omega}_{ib}^b$, the error-contaminated body angular rate. For notational brevity, we still use $\mathbf{C}_{b(t)}^{b(0)}$ instead of $\tilde{\mathbf{C}}_{b(t)}^{b(0)}$ in the sequel.

Substituting (21), the last integral in (15)

$$\begin{aligned}
\int_0^t \mathbf{C}_{b(t)}^{b(0)} \left( \mathbf{f}^b - \mathbf{b}_a \right) dt &= \sum_{k=0}^{M-1} \int_{t_k}^{t_{k+1}} \mathbf{C}_{b(t)}^{b(0)} \left( \mathbf{f}^b - \mathbf{b}_a \right) dt = \sum_{k=0}^{M-1} \mathbf{C}_{b(t_k)}^{b(0)} \int_{t_k}^{t_{k+1}} \mathbf{C}_{b(t)}^{b(t_k)} \left( \mathbf{f}^b - \mathbf{b}_a \right) dt \\
&\approx \sum_{k=0}^{M-1} \left( \mathbf{C}_{b(t_k)}^{b(0)} - kT\mathbf{b}_g \times \right) \cdot \int_{t_k}^{t_{k+1}} \left( \mathbf{I} + \left( \int_{t_k}^{t} \boldsymbol{\omega}_{ib}^b - \mathbf{b}_g d\tau \right) \times \right) \left( \mathbf{f}^b - \mathbf{b}_a \right) dt
\end{aligned} \tag{22}$$

where the incremental integral above can be approximated using the two-sample correction by (see Appendix)

$$\begin{aligned}
&\int_{t_k}^{t_{k+1}}\left(\mathbf{I}+\left(\int_{t_k}^{t}\boldsymbol{\omega}_{ib}^{b}-\mathbf{b}_g d\tau\right)\times\right)\left(\mathbf{f}^b-\mathbf{b}_a\right)dt \\
&=\int_{t_k}^{t_{k+1}}\left(\mathbf{I}+\left(\int_{t_k}^{t}\boldsymbol{\omega}_{ib}^{b}d\tau\right)\times\right)\mathbf{f}^b dt-\int_{t_k}^{t_{k+1}}\left(\mathbf{I}+\left(\int_{t_k}^{t}\boldsymbol{\omega}_{ib}^{b}d\tau\right)\times\right)dt\,\mathbf{b}_a-\mathbf{b}_g\times\int_{t_k}^{t_{k+1}}\left(\mathbf{f}^b-\mathbf{b}_a\right)dt \\
&=\Delta\mathbf{v}_1+\Delta\mathbf{v}_2+\frac{1}{2}\left(\Delta\boldsymbol{\theta}_1+\Delta\boldsymbol{\theta}_2\right)\times\left(\Delta\mathbf{v}_1+\Delta\mathbf{v}_2\right)+\frac{2}{3}\left(\Delta\boldsymbol{\theta}_1\times\Delta\mathbf{v}_2+\Delta\mathbf{v}_1\times\Delta\boldsymbol{\theta}_2\right) \\
&\quad -T\left[\mathbf{I}+\frac{1}{6}\left(5\Delta\boldsymbol{\theta}_1+\Delta\boldsymbol{\theta}_2\right)\times\right]\mathbf{b}_a+\left(\Delta\mathbf{v}_1+\Delta\mathbf{v}_2-T\mathbf{b}_a\right)\times\mathbf{b}_g
\end{aligned} \tag{23}$$

where $\Delta\mathbf{v}_1, \Delta\mathbf{v}_2$ are the first and the second samples of the accelerometer-measured incremental velocity and $\Delta\boldsymbol{\theta}_1, \Delta\boldsymbol{\theta}_2$ are the first and the second samples of the gyroscope-measured incremental angle, respectively, during the update interval $[t_k \quad t_{k+1}]$. In contrast to $\boldsymbol{\omega}_{in}^{n}$, the body rate $\boldsymbol{\omega}_{ib}^{b}$ is fast changing and has to be handled by special treatment as such. This technique is a common practice in the inertial navigation community [1, 16, 17]. Substituting (23) into (22) and neglecting those products of inertial sensor biases higher than one order, we obtain

$$\begin{aligned}
\int_0^t \mathbf{C}_{b(t)}^{b(0)}\left(\mathbf{f}^b-\mathbf{b}_a\right)dt &\approx \sum_{k=0}^{M-1}\mathbf{C}_{b(t_k)}^{b(0)}\left[\Delta\mathbf{v}_1+\Delta\mathbf{v}_2+\frac{1}{2}\left(\Delta\boldsymbol{\theta}_1+\Delta\boldsymbol{\theta}_2\right)\times\left(\Delta\mathbf{v}_1+\Delta\mathbf{v}_2\right)+\frac{2}{3}\left(\Delta\boldsymbol{\theta}_1\times\Delta\mathbf{v}_2+\Delta\mathbf{v}_1\times\Delta\boldsymbol{\theta}_2\right)\right] \\
&\quad -\sum_{k=0}^{M-1}T\mathbf{C}_{b(t_k)}^{b(0)}\left[\mathbf{I}+\frac{1}{6}\left(5\Delta\boldsymbol{\theta}_1+\Delta\boldsymbol{\theta}_2\right)\times\right]\mathbf{b}_a \\
&\quad +\sum_{k=0}^{M-1}\left\{\mathbf{C}_{b(t_k)}^{b(0)}\left(\Delta\mathbf{v}_1+\Delta\mathbf{v}_2\right)\times\mathbf{b}_g\right. \\
&\quad \left.-kT\mathbf{b}_g\times\left[\Delta\mathbf{v}_1+\Delta\mathbf{v}_2+\frac{1}{2}\left(\Delta\boldsymbol{\theta}_1+\Delta\boldsymbol{\theta}_2\right)\times\left(\Delta\mathbf{v}_1+\Delta\mathbf{v}_2\right)+\frac{2}{3}\left(\Delta\boldsymbol{\theta}_1\times\Delta\mathbf{v}_2+\Delta\mathbf{v}_1\times\Delta\boldsymbol{\theta}_2\right)\right]\right\} \\
&\triangleq \boldsymbol{\alpha}_M+\boldsymbol{\chi}_M\mathbf{b}_a+\boldsymbol{\lambda}_M\mathbf{b}_g
\end{aligned} \tag{24}$$

where the symbols are defined as

$$\begin{aligned}
\boldsymbol{\alpha}_M &= \sum_{k=0}^{M-1} \mathbf{C}_{b(t_k)}^{b(0)} \Delta\mathbf{v} \\
\boldsymbol{\chi}_M &= -\sum_{k=0}^{M-1} T\mathbf{C}_{b(t_k)}^{b(0)} \left[ \mathbf{I} + \frac{1}{6}(5\Delta\boldsymbol{\theta}_1 + \Delta\boldsymbol{\theta}_2)\times \right] \\
\boldsymbol{\lambda}_M &= \sum_{k=0}^{M-1} \left[ \mathbf{C}_{b(t_k)}^{b(0)} \left(\Delta\mathbf{v}_1 + \Delta\mathbf{v}_2\right)\times + kT\Delta\mathbf{v}\times \right] \\
\Delta\mathbf{v} &= \Delta\mathbf{v}_1 + \Delta\mathbf{v}_2 + \frac{1}{2}\left(\Delta\boldsymbol{\theta}_1 + \Delta\boldsymbol{\theta}_2\right)\times\left(\Delta\mathbf{v}_1 + \Delta\mathbf{v}_2\right) + \frac{2}{3}\left(\Delta\boldsymbol{\theta}_1 \times \Delta\mathbf{v}_2 + \Delta\mathbf{v}_1 \times \Delta\boldsymbol{\theta}_2\right)
\end{aligned} \tag{25}$$

Similarly with (21) and neglecting the multiplication term of unknowns, the remaining term of (15)

$$\left(\mathbf{C}_{b(t_M)}^{b(0)} \boldsymbol{\omega}_{ib}^b \times - \boldsymbol{\omega}_{ib}^b(0)\times\right)\mathbf{l}^b \triangleq \boldsymbol{\gamma}_M \mathbf{l}^b \tag{26}$$

where the new symbol $\boldsymbol{\gamma}_M$ is obviously defined.

With (24) and (26), the integration equation (15) is rewritten as

$$\boldsymbol{\beta}_M = \mathbf{C}_b^n(0)\left(\boldsymbol{\alpha}_M + \boldsymbol{\chi}_M \mathbf{b}_a + \boldsymbol{\lambda}_M \mathbf{b}_g + \boldsymbol{\gamma}_M \mathbf{l}^b\right), \quad M = 0,1,2,\ldots \tag{27}$$

which consists of a number of equalities at all times. The unknown constant parameters here include the initial attitude matrix $\mathbf{C}_b^n(0)$, the accelerometer bias $\mathbf{b}_a$, the gyroscope bias $\mathbf{b}_g$ and the lever arm $\mathbf{l}^b$. As regards (25) and (26), the coeffiicents $\boldsymbol{\alpha}_M$, $\boldsymbol{\chi}_M$, $\boldsymbol{\lambda}_M$ and $\boldsymbol{\gamma}_M$ are all time-varying quantities computed using the raw gyroscope/accelerometer outputs. The equalities in (27) are supposed to be used to form the objective function for optimization in next section, and should be properly weighted for an optimal parameter estimation. The simplest way is to give equal weights to all equalities. However, the equality for different $M$ has quite different error characteristics, because it is an integration over a time interval with varying length. Here we construct a new set of equalities by even-interval time difference

$$\boldsymbol{\beta}_M - \boldsymbol{\beta}_{M-\nabla} = \mathbf{C}_b^n(0)\left[\boldsymbol{\alpha}_M - \boldsymbol{\alpha}_{M-\nabla} + (\boldsymbol{\chi}_M - \boldsymbol{\chi}_{M-\nabla})\mathbf{b}_a + (\boldsymbol{\lambda}_M - \boldsymbol{\lambda}_{M-\nabla})\mathbf{b}_g + (\boldsymbol{\gamma}_M - \boldsymbol{\gamma}_{M-\nabla})\mathbf{l}^b\right], \quad M = \nabla, \nabla+1, \nabla+2, \ldots \tag{28}$$

where the length of the time difference interval is $(\nabla+1)T$. For notational brevity, we re-express the above equation (28) as

$$\boldsymbol{\beta} = \mathbf{C}_b^n(0)\left(\boldsymbol{\alpha} + \boldsymbol{\chi}\mathbf{b}_a + \boldsymbol{\lambda}\mathbf{b}_g + \boldsymbol{\gamma}\mathbf{l}^b\right), \quad M = \nabla, \nabla+1, \nabla+2, \ldots \tag{29}$$

where the coefficients $\boldsymbol{\alpha} \triangleq \boldsymbol{\alpha}_M - \boldsymbol{\alpha}_{M-\nabla}$, $\boldsymbol{\beta} \triangleq \boldsymbol{\beta}_M - \boldsymbol{\beta}_{M-\nabla}$, $\boldsymbol{\chi} \triangleq \boldsymbol{\chi}_M - \boldsymbol{\chi}_{M-\nabla}$, $\boldsymbol{\lambda} \triangleq \boldsymbol{\lambda}_M - \boldsymbol{\lambda}_{M-\nabla}$ and

$\boldsymbol{\gamma} \triangleq \boldsymbol{\gamma}_M - \boldsymbol{\gamma}_{M-\nabla}$.

## IV. Joint Attitude and Parameter Estimation

Next we will pose (28) as a constrained minimization problem by use of the unit quaternion to replace the attitude matrix as in [10, 12, 19]. Specifically, the four-element unit quaternion $\mathbf{q} = [s \quad \boldsymbol{\eta}^T]^T$, where $s$ is the scalar part and $\boldsymbol{\eta}$ is the vector part, is used to encode the initial body attitude matrix $\mathbf{C}_n^b(0)$. The two rotation parameters are related by (note a sign typo in (9) of [12])

$$\mathbf{C}_n^b(0) = \left(s^2 - \boldsymbol{\eta}^T\boldsymbol{\eta}\right)\mathbf{I} + 2\boldsymbol{\eta}\boldsymbol{\eta}^T - 2s(\boldsymbol{\eta}\times) \tag{30}$$

Define the quaternion multiplication matrices by

$$[\overset{+}{\mathbf{q}}] \triangleq \begin{bmatrix} s & -\boldsymbol{\eta}^T \\ \boldsymbol{\eta} & sI + (\boldsymbol{\eta}\times) \end{bmatrix}, \quad [\overset{-}{\mathbf{q}}] \triangleq \begin{bmatrix} s & -\boldsymbol{\eta}^T \\ \boldsymbol{\eta} & sI - (\boldsymbol{\eta}\times) \end{bmatrix} \tag{31}$$

such that $\mathbf{q} \circ \mathbf{p} = [\overset{+}{\mathbf{q}}]\mathbf{p} = [\overset{-}{\mathbf{p}}]\mathbf{q}$ for any two quaternions $\mathbf{p}$ and $\mathbf{q}$. The operator $\circ$ means the quaternion multiplication [20]. It is easy to verify the following quaternion equalities as

$$\begin{aligned} &[\overset{+}{\mathbf{q}}][\overset{-}{\mathbf{p}}] = [\overset{-}{\mathbf{p}}][\overset{+}{\mathbf{q}}], \quad [\overset{+}{\mathbf{q}}][\overset{+}{\mathbf{p}}] = [\overset{+}{\mathbf{q} \circ \mathbf{p}}] \\ &[\overset{+}{\mathbf{q}}]^T = [\overset{+}{\mathbf{q}^*}], \quad [\overset{-}{\mathbf{q}}]^T = [\overset{-}{\mathbf{q}^*}] \end{aligned} \tag{32}$$

Then (28) is equivalent to

$$0 = \left([\overset{-}{\boldsymbol{\alpha}}] - [\overset{+}{\boldsymbol{\beta}}]\right)\mathbf{q} + [\overset{+}{\mathbf{q}}]\left(\boldsymbol{\chi}\mathbf{b}_a + \boldsymbol{\lambda}\mathbf{b}_g + \boldsymbol{\gamma}\mathbf{l}^b\right) \triangleq \boldsymbol{\pi} \tag{33}$$

From here on, a vector is used interchangeably with a vector quaternion, namely, a quaternion with zero scalar part and the vector part being the vector. The problem can be posed as a unit quaternion-constrained optimization

$$\min_{\mathbf{q},\mathbf{l}^b,\mathbf{b}_a,\mathbf{b}_g} \sum_M \boldsymbol{\pi}^T\boldsymbol{\pi}, \text{ subject to } \mathbf{q}^T\mathbf{q} = 1 \tag{34}$$

Ignoring the sensor biases and the lever arm, it is reduced to the pure attitude alignment in [10, 12], i.e.,

$$\min_{\mathbf{q}} \sum_M \mathbf{q}^T \left( [\bar{\boldsymbol{\alpha}}] - [\overset{+}{\boldsymbol{\beta}}] \right)^T \left( [\bar{\boldsymbol{\alpha}}] - [\overset{+}{\boldsymbol{\beta}}] \right) \mathbf{q}, \text{ subject to } \mathbf{q}^T \mathbf{q} = 1 \tag{35}$$

which can be exactly solved using the eigenvalue problem [21]. But in general, (34) has no analytical solution and we are obliged to seek an approximate solution.

*A. Batch Algorithm*

This paper chooses the iterative Newton-Lagrange method to attack the nonlinearly constrained optimization problem (34), because it has proved to be very effective in practice and is the basis of some of the best software for solving both small and large constrained optimization problems [22].

The Lagangian for the problem (34) is defined as

$$L(\mathbf{x}, \mu) = \sum_M \boldsymbol{\pi}^T \boldsymbol{\pi} - \mu \left( \mathbf{q}^T \mathbf{q} - 1 \right) \tag{36}$$

where $\mathbf{x} \triangleq \begin{bmatrix} \mathbf{q}^T & \mathbf{b}_a^T & \mathbf{b}_g^T & \mathbf{l}^{bT} \end{bmatrix}^T$ and $\mu$ is the Lagrange multiplier. The iterative step starting from $(\mathbf{x}_k, \mu_k)$ is given by

$$\begin{bmatrix} \mathbf{x}_{k+1} \\ \mu_{k+1} \end{bmatrix} = \begin{bmatrix} \mathbf{x}_k \\ \mu_k \end{bmatrix} + \begin{bmatrix} \Delta \mathbf{x} \\ \Delta \mu \end{bmatrix} \tag{37}$$

where $(\Delta \mathbf{x}, \Delta \mu)$ satisfies the following system

$$\begin{bmatrix} \nabla^2_{\mathbf{xx}} L(\mathbf{x}_k, \mu_k) & \nabla^2_{\mathbf{x}\mu} L(\mathbf{x}_k, \mu_k) \\ -\nabla^2_{\mathbf{x}\mu} L(\mathbf{x}_k, \mu_k)^T & 0 \end{bmatrix} \begin{bmatrix} \Delta \mathbf{x} \\ \Delta \mu \end{bmatrix} = - \begin{bmatrix} \nabla_{\mathbf{x}} L(\mathbf{x}_k, \mu_k) \\ \mathbf{q}_k^T \mathbf{q}_k - 1 \end{bmatrix} \tag{38}$$

The coefficient matrix is nonsingular and the iteration is quadratically convergent under moderate assumptions provided that the starting point $\mathbf{x}_0$ is close enough to the true value [22]. For the problem under investigation, a good starting point is available as discussed in *Remark 2.2*.

With (36), we can obtain the first and second order derivatives as

$$\begin{aligned} \nabla_{\mathbf{x}} L(\mathbf{x}_k, \mu_k) &= 2 \sum_M \mathbf{J} - 2\mu \begin{bmatrix} \mathbf{q}^T & \mathbf{0}_{1\times3} & \mathbf{0}_{1\times3} & \mathbf{0}_{1\times3} \end{bmatrix}^T \\ \nabla^2_{\mathbf{x}\mu} L(\mathbf{x}_k, \mu_k) &= -2 \begin{bmatrix} \mathbf{q}^T & \mathbf{0}_{1\times3} & \mathbf{0}_{1\times3} & \mathbf{0}_{1\times3} \end{bmatrix}^T \\ \nabla^2_{\mathbf{xx}} L(\mathbf{x}_k, \mu_k) &= 2 \sum_M \mathbf{H} - 2\mu \, \mathrm{diag}\left( \begin{bmatrix} \mathbf{1}_{1\times4} & \mathbf{0}_{1\times9} \end{bmatrix}^T \right) \end{aligned} \tag{39}$$

where $\mathbf{J} \triangleq \begin{bmatrix} \mathbf{J}_1^T & \mathbf{J}_2^T & \mathbf{J}_3^T & \mathbf{J}_4^T \end{bmatrix}^T$ is the Jacobian matrix of $\boldsymbol{\pi}^T\boldsymbol{\pi}$ with

$$
\begin{aligned}
\mathbf{J}_1 &= \left(\overset{-}{[\boldsymbol{\alpha}]}-\overset{+}{[\boldsymbol{\beta}]}\right)^T \left(\overset{-}{[\boldsymbol{\alpha}]}-\overset{+}{[\boldsymbol{\beta}]}\right)\mathbf{q} + \left(\overset{-}{[\boldsymbol{\alpha}]}-\overset{+}{[\boldsymbol{\beta}]}\right)^T \overset{+}{[\mathbf{q}]}\left(\boldsymbol{\chi}\mathbf{b}_a + \boldsymbol{\lambda}\mathbf{b}_g + \boldsymbol{\gamma}\mathbf{l}^b\right) + \overset{-}{\left[\boldsymbol{\chi}\mathbf{b}_a + \boldsymbol{\lambda}\mathbf{b}_g + \boldsymbol{\gamma}\mathbf{l}^b\right]}^T \left(\overset{-}{[\boldsymbol{\alpha}]}-\overset{+}{[\boldsymbol{\beta}]}\right)\mathbf{q} \\
\mathbf{J}_2 &= \boldsymbol{\chi}^T \left\{ \overset{+}{[\mathbf{q}]}^T \left(\overset{-}{[\boldsymbol{\alpha}]}-\overset{+}{[\boldsymbol{\beta}]}\right)\mathbf{q} + \left(\boldsymbol{\chi}\mathbf{b}_a + \boldsymbol{\lambda}\mathbf{b}_g + \boldsymbol{\gamma}\mathbf{l}^b\right) \right\} \\
\mathbf{J}_3 &= \boldsymbol{\lambda}^T \left\{ \overset{+}{[\mathbf{q}]}^T \left(\overset{-}{[\boldsymbol{\alpha}]}-\overset{+}{[\boldsymbol{\beta}]}\right)\mathbf{q} + \left(\boldsymbol{\chi}\mathbf{b}_a + \boldsymbol{\lambda}\mathbf{b}_g + \boldsymbol{\gamma}\mathbf{l}^b\right) \right\} \\
\mathbf{J}_4 &= \boldsymbol{\gamma}^T \left\{ \overset{+}{[\mathbf{q}]}^T \left(\overset{-}{[\boldsymbol{\alpha}]}-\overset{+}{[\boldsymbol{\beta}]}\right)\mathbf{q} + \left(\boldsymbol{\chi}\mathbf{b}_a + \boldsymbol{\lambda}\mathbf{b}_g + \boldsymbol{\gamma}\mathbf{l}^b\right) \right\}
\end{aligned}
\tag{40}
$$

and $\mathbf{H} \triangleq \begin{bmatrix} \mathbf{H}_{11} & \mathbf{H}_{12} & \mathbf{H}_{13} & \mathbf{H}_{14} \\ \mathbf{H}_{12}^T & \mathbf{H}_{22} & \mathbf{H}_{23} & \mathbf{H}_{24} \\ \mathbf{H}_{13}^T & \mathbf{H}_{23}^T & \mathbf{H}_{33} & \mathbf{H}_{34} \\ \mathbf{H}_{14}^T & \mathbf{H}_{24}^T & \mathbf{H}_{34}^T & \mathbf{H}_{44} \end{bmatrix}$ is the Hessian matrix of $\boldsymbol{\pi}^T\boldsymbol{\pi}$ with

$$
\begin{aligned}
\mathbf{H}_{11} &= \left(\overset{-}{[\boldsymbol{\alpha}]}-\overset{+}{[\boldsymbol{\beta}]}\right)^T \left(\overset{-}{[\boldsymbol{\alpha}]}-\overset{+}{[\boldsymbol{\beta}]}\right) + \left(\overset{-}{[\boldsymbol{\alpha}]}-\overset{+}{[\boldsymbol{\beta}]}\right)^T \overset{-}{\left[\boldsymbol{\chi}\mathbf{b}_a + \boldsymbol{\lambda}\mathbf{b}_g + \boldsymbol{\gamma}\mathbf{l}^b\right]} + \overset{-}{\left[\boldsymbol{\chi}\mathbf{b}_a + \boldsymbol{\lambda}\mathbf{b}_g + \boldsymbol{\gamma}\mathbf{l}^b\right]}^T \left(\overset{-}{[\boldsymbol{\alpha}]}-\overset{+}{[\boldsymbol{\beta}]}\right) \\
\mathbf{H}_{12} &= \left\{ \left(\overset{-}{[\boldsymbol{\alpha}]}-\overset{+}{[\boldsymbol{\beta}]}\right)^T \overset{+}{[\mathbf{q}]} - \overset{+}{\left[\left(\overset{-}{[\boldsymbol{\alpha}]}-\overset{+}{[\boldsymbol{\beta}]}\right)\mathbf{q}\right]} \right\} \boldsymbol{\chi} \\
\mathbf{H}_{13} &= \left\{ \left(\overset{-}{[\boldsymbol{\alpha}]}-\overset{+}{[\boldsymbol{\beta}]}\right)^T \overset{+}{[\mathbf{q}]} - \overset{+}{\left[\left(\overset{-}{[\boldsymbol{\alpha}]}-\overset{+}{[\boldsymbol{\beta}]}\right)\mathbf{q}\right]} \right\} \boldsymbol{\lambda} \\
\mathbf{H}_{14} &= \left\{ \left(\overset{-}{[\boldsymbol{\alpha}]}-\overset{+}{[\boldsymbol{\beta}]}\right)^T \overset{+}{[\mathbf{q}]} - \overset{+}{\left[\left(\overset{-}{[\boldsymbol{\alpha}]}-\overset{+}{[\boldsymbol{\beta}]}\right)\mathbf{q}\right]} \right\} \boldsymbol{\gamma} \\
\mathbf{H}_{22} &= \boldsymbol{\chi}^T\boldsymbol{\chi}, \mathbf{H}_{23} = \boldsymbol{\chi}^T\boldsymbol{\lambda}, \mathbf{H}_{24} = \boldsymbol{\chi}^T\boldsymbol{\gamma} \\
\mathbf{H}_{33} &= \boldsymbol{\lambda}^T\boldsymbol{\lambda}, \mathbf{H}_{34} = \boldsymbol{\lambda}^T\boldsymbol{\gamma}, \mathbf{H}_{44} = \boldsymbol{\gamma}^T\boldsymbol{\gamma}
\end{aligned}
\tag{41}
$$

In the above development, the quaternion equalities in (32) have been frequently used.

***Batch Algorithm for Joint Attitude and Parameter Estimation (BA-JAPE):***

*For the current time* $t_M$ *,*

*Step 1: Compute* $\boldsymbol{\alpha}$*,* $\boldsymbol{\beta}$*,* $\boldsymbol{\chi}$*,* $\boldsymbol{\lambda}$ *and* $\boldsymbol{\gamma}$ *according to (20), (25) and (26);*

*Step 2: Set* $k=0$ *and start with a good estimate* $\left(\mathbf{x}_k,\mu_k\right)$*;*

*Step 3: Compute the derivatives* $\nabla_{\mathbf{x}}L\left(\mathbf{x}_k,\mu_k\right)$*,* $\nabla^2_{\mathbf{x}\mu}L\left(\mathbf{x}_k,\mu_k\right)^T$ *and* $\nabla^2_{\mathbf{xx}}L\left(\mathbf{x}_k,\mu_k\right)$ *using (39)-(41);*

*Step 4: Solve* $\left(\Delta\mathbf{x},\Delta\mu\right)$ *according to (38);*

*Step 5: Update* $\left(\mathbf{x}_k,\mu_k\right)$ *with (37);*

*Step 6: Set* $k=k+1$ *and go to Step 3 until convergence or maximum iteration is reached;*

*Step 7: Obtain the current attitude from the initial attitude using* (9).

We say that BA-JAPE is a batch algorithm because the derivatives in Step 3 have to be computed at $\left(\mathbf{x}_k,\mu_k\right)$ using all history data for each iteration. In fact, as we will see, the forms of (20), (25)-(26) and (40)-(41) enable a time-recursive algorithm that only uses current data. The main difficulty is how to put the constant parameters, namely $\mathbf{q}$, $\mathbf{b}_a$, $\mathbf{b}_g$ and $\mathbf{l}^b$, out of the two summations $\sum_M \mathbf{J}$ and $\sum_M \mathbf{H}$ in (39) by proper transformations.

*B. Recursive Algorithm*

We now try to manipulate the parameters in (39) as left/right multiplying factors so as to be taken out of the summation symbols. This will offer us a recursive and equivalent form. Specifically, we can derive

$$\mathbf{J}_1=\left(\overset{-}{[\boldsymbol{\alpha}]}-\overset{+}{[\boldsymbol{\beta}]}\right)^T\left(\overset{-}{[\boldsymbol{\alpha}]}-\overset{+}{[\boldsymbol{\beta}]}\right)\mathbf{q}-\overset{+}{[\mathbf{q}]}\left(\overset{-}{[\boldsymbol{\alpha}]}+\overset{+}{[\boldsymbol{\alpha}]}\right)\left(\boldsymbol{\chi}\mathbf{b}_a+\boldsymbol{\lambda}\mathbf{b}_g+\boldsymbol{\gamma}\mathbf{l}^b\right)+2\overset{+}{[\boldsymbol{\beta}]}\overset{+}{[\mathbf{q}]}\left(\boldsymbol{\chi}\mathbf{b}_a+\boldsymbol{\lambda}\mathbf{b}_g+\boldsymbol{\gamma}\mathbf{l}^b\right)\tag{42}$$

The third term does not fulfill the requirement, but the left-multiplying factor $\overset{+}{[\boldsymbol{\beta}]}\overset{+}{[\mathbf{q}]}$ can be decomposed as

$$\overset{+}{[\boldsymbol{\beta}]}\overset{+}{[\mathbf{q}]}=s\overset{+}{[\boldsymbol{\beta}]}+\eta_1\mathbf{J}_{\boldsymbol{\beta}1}+\eta_2\mathbf{J}_{\boldsymbol{\beta}2}+\eta_3\mathbf{J}_{\boldsymbol{\beta}3}\tag{43}$$

$\eta_i$ is the component of the vector part $\boldsymbol{\eta}$ of $\mathbf{q}$ and

$$\mathbf{J}_{\beta_1}=\begin{bmatrix}-\beta_1 & 0 & -\beta_3 & \beta_2\\ 0 & -\beta_1 & \beta_2 & \beta_3\\ \beta_3 & -\beta_2 & -\beta_1 & 0\\ -\beta_2 & -\beta_3 & 0 & -\beta_1\end{bmatrix},\mathbf{J}_{\beta_2}=\begin{bmatrix}-\beta_2 & \beta_3 & 0 & -\beta_1\\ -\beta_3 & -\beta_2 & -\beta_1 & 0\\ 0 & \beta_1 & -\beta_2 & \beta_3\\ \beta_1 & 0 & -\beta_3 & -\beta_2\end{bmatrix},\mathbf{J}_{\beta_3}=\begin{bmatrix}-\beta_3 & -\beta_2 & \beta_1 & 0\\ \beta_2 & -\beta_3 & 0 & -\beta_1\\ -\beta_1 & 0 & -\beta_3 & -\beta_2\\ 0 & \beta_1 & \beta_2 & -\beta_3\end{bmatrix}\tag{44}$$

where $\beta_i$ is the $i^{\text{th}}$ component of $\boldsymbol{\beta}$. Now substituting (42)-(44), we see that the summation term related to $\mathbf{J}_1$ in (39) can be expressed as

$$\begin{aligned}\sum_M \mathbf{J}_1 = &\left\{\sum_M \left(\overset{-}{[\boldsymbol{\alpha}]}-\overset{+}{[\boldsymbol{\beta}]}\right)^T \left(\overset{-}{[\boldsymbol{\alpha}]}-\overset{+}{[\boldsymbol{\beta}]}\right)\right\}\mathbf{q} - \overset{+}{[\mathbf{q}]}\left\{\sum_M \left(\overset{-}{[\boldsymbol{\alpha}]}+\overset{+}{[\boldsymbol{\alpha}]}\right)[\boldsymbol{\chi} \quad \boldsymbol{\lambda} \quad \boldsymbol{\gamma}]\right\}\begin{bmatrix}\mathbf{b}_a^T & \mathbf{b}_g^T & \mathbf{l}^{bT}\end{bmatrix}^T \\ &+2\left\{s\sum_M \overset{+}{[\boldsymbol{\beta}]}[\boldsymbol{\chi} \quad \boldsymbol{\lambda} \quad \boldsymbol{\gamma}] + \sum_{i=1}^{3}\left\{\eta_i \sum_M \mathbf{J}_{\beta_i}[\boldsymbol{\chi} \quad \boldsymbol{\lambda} \quad \boldsymbol{\gamma}]\right\}\right\}\begin{bmatrix}\mathbf{b}_a^T & \mathbf{b}_g^T & \mathbf{l}^{bT}\end{bmatrix}^T\end{aligned} \tag{45}$$

For $\mathbf{J}_2$, we have

$$\mathbf{J}_2 = \boldsymbol{\chi}^T\left\{\boldsymbol{\alpha} - \overset{+}{[\mathbf{q}^*]}\overset{+}{[\boldsymbol{\beta}]}\mathbf{q} + \left(\boldsymbol{\chi}\mathbf{b}_a + \boldsymbol{\lambda}\mathbf{b}_g + \boldsymbol{\gamma}\mathbf{l}^b\right)\right\} \tag{46}$$

The second term in the bracket $\overset{+}{[\mathbf{q}^*]}\overset{+}{[\boldsymbol{\beta}]}\mathbf{q}$ can be decomposed as

$$\overset{+}{[\mathbf{q}^*]}\overset{+}{[\boldsymbol{\beta}]}\mathbf{q} = \beta_1\mathbf{J}_{q_1} + \beta_2\mathbf{J}_{q_2} + \beta_3\mathbf{J}_{q_3} \tag{47}$$

where

$$\mathbf{J}_{q_1} = \begin{bmatrix} 0 \\ s^2+\eta_1^2-\eta_2^2-\eta_3^2 \\ 2\eta_1\eta_2 - 2s\eta_3 \\ 2s\eta_2 + 2\eta_1\eta_3 \end{bmatrix}, \mathbf{J}_{q_2} = \begin{bmatrix} 0 \\ 2\eta_1\eta_2 + 2s\eta_3 \\ s^2-\eta_1^2+\eta_2^2-\eta_3^2 \\ -2s\eta_1 + 2\eta_2\eta_3 \end{bmatrix}, \mathbf{J}_{q_3} = \begin{bmatrix} 0 \\ -2s\eta_2 + 2\eta_1\eta_3 \\ 2s\eta_1 + 2\eta_2\eta_3 \\ s^2-\eta_1^2-\eta_2^2+\eta_3^2 \end{bmatrix} \tag{48}$$

Substituting (46)-(48), the summation term related to $\mathbf{J}_2$ in (39) is expressed as

$$\sum_M \mathbf{J}_2 = \sum_M \boldsymbol{\chi}^T\boldsymbol{\alpha} - \sum_{i=1}^{3}\left\{\left(\sum_M \beta_i\boldsymbol{\chi}^T\right)\mathbf{J}_{q_i}\right\} + \left\{\sum_M \boldsymbol{\chi}^T[\boldsymbol{\chi} \quad \boldsymbol{\lambda} \quad \boldsymbol{\gamma}]\right\}\begin{bmatrix}\mathbf{b}_a^T & \mathbf{b}_g^T & \mathbf{l}^{bT}\end{bmatrix}^T \tag{49}$$

Similar treatment with $\mathbf{J}_2$ also applies to $\mathbf{J}_3$ and $\mathbf{J}_4$, because they are all of the same structure (cf. (40)).

As for $\mathbf{H}_{11}$, we obtain

$$\mathbf{H}_{11} = \left(\overset{-}{[\boldsymbol{\alpha}]}-\overset{+}{[\boldsymbol{\beta}]}\right)^T\left(\overset{-}{[\boldsymbol{\alpha}]}-\overset{+}{[\boldsymbol{\beta}]}\right) + \left(\overset{-}{[\boldsymbol{\alpha}]}-\overset{+}{[\boldsymbol{\beta}]}\right)^T\overset{-}{\left[\boldsymbol{\chi}\mathbf{b}_a + \boldsymbol{\lambda}\mathbf{b}_g + \boldsymbol{\gamma}\mathbf{l}^b\right]} + \overset{-}{\left[\boldsymbol{\chi}\mathbf{b}_a + \boldsymbol{\lambda}\mathbf{b}_g + \boldsymbol{\gamma}\mathbf{l}^b\right]}^T\left(\overset{-}{[\boldsymbol{\alpha}]}-\overset{+}{[\boldsymbol{\beta}]}\right) \tag{50}$$

The last two terms do not fulfill the requirement but they are transposition of each other. Let us pay attention to the components of the second term only. Specifically, $\overset{-}{[\boldsymbol{\chi}\mathbf{b}_a]}$ can be decomposed as

$$\left[\bar{\chi \mathbf{b}_a}\right]=\left[b_{a1}\bar{\chi}_1+b_{a2}\bar{\chi}_2+b_{a3}\bar{\chi}_3\right]=b_{a1}\left[\bar{\chi}_1\right]+b_{a2}\left[\bar{\chi}_2\right]+b_{a3}\left[\bar{\chi}_3\right] \tag{51}$$

where $\chi_i$ is the $i$th column of $\chi$ and $b_{ai}$ is the $i$th component of $\mathbf{b}_a$. The other components $\left[\bar{\lambda \mathbf{b}_g}\right]$ and $\left[\bar{\gamma \mathbf{l}^b}\right]$ can be treated as such.

Using the quaternion identities in (32), $\mathbf{H}_{12}$ can be developed as

$$\mathbf{H}_{12}=-\left\{\overset{+}{[\mathbf{q}]}\overset{-}{[\alpha]}-\overset{+}{[\beta]}\overset{+}{[\mathbf{q}]}+\overset{+}{[\mathbf{q}\circ\alpha-\beta\circ\mathbf{q}]}\right\}\chi=\left\{2\overset{+}{[\beta]}\overset{+}{[\mathbf{q}]}-\overset{+}{[\mathbf{q}]}\left(\overset{-}{[\alpha]}+\overset{+}{[\alpha]}\right)\right\}\chi \tag{52}$$

The first term in the bracket takes the same decomposition as in (43). Similar treatment applies to $\mathbf{H}_{13}$ and $\mathbf{H}_{14}$ as well.

Other terms of the Hessian matrix $\mathbf{H}$ (cf. (41)) are independent of the parameters and need no special treatment.

***Recursive Algorithm for Joint Attitude and Parameter Estimation (RA-JAPE):***

*For the current time* $t_M$,

*Step 1: Incrementally compute* $\alpha$, $\beta$, $\chi$, $\lambda$ *and* $\gamma$ *according to (20), (25) and (26). Only their previous values at time* $t_{M-1}$ *are used;*

*Step 2: Incrementally compute the two summation terms* $\sum_M \mathbf{J}$ and $\sum_M \mathbf{H}$ *using* (42)-(52). *Only their previous values at time* $t_{M-1}$ *are used;*

*Step 3: Set* $k=0$ *and start with a good estimate* $(\mathbf{x}_k,\mu_k)$. *The initial attitude can be acquired by solving (35) with* $\alpha$, $\beta$;

*Step 4: Calculate the derivatives* $\nabla_{\mathbf{x}}L(\mathbf{x}_k,\mu_k)$, $\nabla^2_{\mathbf{x}\mu}L(\mathbf{x}_k,\mu_k)^T$ *and* $\nabla^2_{\mathbf{xx}}L(\mathbf{x}_k,\mu_k)$ *according to (39);*

*Step 5: Solve* $(\Delta\mathbf{x},\Delta\mu)$ *according to (38);*

*Step 6: Update* $(\mathbf{x}_k,\mu_k)$ *with (37);*

*Step 7: Set* $k = k + 1$ *and go to Step 4 until convergence or maximum iteration is reached;*

*Step 8: Obtain the current attitude from the initial attitude using* (9).

The RA-JAPE algorithm is derived from the Newton-Lagrange method that is generally sensitive to the starting point, so we should be cautious about starting the algorithm with a good initialization. Fortunately, the RA-JAPE can initialize the attitude parameter with attitude-only estimation (cf. (35)) simply by neglecting gyroscope/accelerometer biases and the GNSS lever arm, which is equivalent to the recursive attitude alignment algorithm in [10]. The computation involved is relatively negligible. Other parameters start from zero. Note that the initial attitude estimate is intrinsic to the RA-JAPE and readily available at any time instant. Figure 1 presents the flowchart of the RA-JAPE algorithm. We find by simulation that if the initial attitude were good enough, the RA-JAPE algorithm would converge well within five iterations. Thus the maximum iteration is set to five in the algorithm.

## V. Simulation Results

We design scenarios with oscillating attitude and translation to simulate large motions. The INS is assumed to be located at medium latitude 30° and equipped with a triad of gyroscopes (drift $0.01°/h$, noise $0.1°/h/\sqrt{\mathrm{Hz}}$) and accelerometers (bias $50\mu g$, noise $5\mu g/\sqrt{\mathrm{Hz}}$) at a sampling rate 100 *Hz*. The GPS antenna is displaced from the INS by the lever arm $\mathbf{l}^b = \begin{bmatrix} 1 & 2 & 1.5 \end{bmatrix}^T$ in meters and the GPS receiver outputs the aided velocity and position at 50Hz. The incremental update interval $T = 0.02s$ and the time difference interval in (28) is set to 1 second. Figures 2-4 show the true angular velocity and acceleration profiles experienced by the INS, and the true velocity profile experienced by the GPS antenna, for the first 100 seconds.

To validate the above development and algorithm implementation, we first examine an ideal case without INS and GPS measurement noises. Figure 5 plots the attitude estimate error of the RA-JAPE algorithm, overlapped by the initial attitude estimate error of the pure attitude algorithm [10]. The negative effect of neglecting the lever arm, slow convergence and low accuracy, is apparent in Fig. 5. After accumulating data for about 30 seconds, a good initial heading angle is reached to the accuracy of 8 degree (in contrast to sub degree for the other two angles) and the attitude begins to converge, ending with all three angle errors below 0.001 degree at 300s. Estimates of the sensor biases and the GPS lever arm are given in Figs. 6-8, respectively. The accelerometer bias and the lever arm are estimated to the accuracy of a few micro-g and 0.1 millimeter, respectively. The gyroscope bias estimate result is anticipated because the first-order approximation (21) is insufficient for large three-axis attitude motions. Figure 9 presents a demonstrating behavior of the RA-JAPE algorithm at 300s for ten iterations, where all estimates converge well within five iterations.

The RA-JAPE algorithm is then evaluated by 50 Monte Carlo runs. White velocity noise (standard variance 0.02 *m/s*) and white position noise (standard variance 0.2 *m*) are simulated in GPS measurements. Figure 10 plots the mean and standard variance of initial attitude errors obtained from the pure attitude algorithm and Figure 11 plots the mean and standard variance of current attitude errors of the RA-JAPE algorithm. The three angle errors ($1\sigma$) in Fig. 11 reduce to below $\begin{bmatrix} 0.001 \pm 0.002 & 0.001 \pm 0.003 & -0.0005 \pm 0.002 \end{bmatrix}$ in degree within 300s, in contrast to the initial angle errors at $\begin{bmatrix} 0.01 \pm 0.005 & 0.02 \pm 0.02 & 0.005 \pm 0.005 \end{bmatrix}$ in degree in Fig. 10. Figures 12-13 give the results for the accelerometer bias and the gyroscope bias, respectively. The accelerometer bias estimate stabilizes at $\begin{bmatrix} 52 \pm 18 & 52 \pm 13 & 52 \pm 13 \end{bmatrix}$ in micro-g, which is very close in mean to the true value with tens of micro-g in standard variance. In contrast, the gyroscope bias estimate apparently undulates near zero. Figure 14 plots the estimate errors of the GPS lever arm across Monte Carlo runs. The lever arm errors ($1\sigma$) reduce to $\begin{bmatrix} 0.1 \pm 1.9 & 0.2 \pm 1.8 & 0.1 \pm 1.6 \end{bmatrix}$ in millimeter. The significant accuracy gain in attitude (Fig. 11 vs. Fig. 10) is mainly owed to effective estimation of the lever arm and the accelerometer bias. To further evaluate the RA-JAPE algorithm, the objective function values (34) both at the estimate and at the true value of $\mathbf{x}$ for each run are plotted in Fig. 15. We see that the objective function has been effectively minimized resulting in consistently smaller value than that at the truth. For example, in the 50$^{th}$ run the objective function value is 29.9279 at the estimate versus 29.9287 at the truth.

If only properly initialized, the Kalman filtering method [2] would be applicable in the on-the-move situation. We implement an error-state (or indirect) EKF here as a comparison, with the initial attitude enabled by the attitude-only estimation (cf. (35)). Specifically, the EKF does not start to work until 30s when a good initial attitude is provided. This time period of 30 seconds is carefully selected to make sure the EKF works properly, and the EKF's parameters (such as the initial state covariance) are tuned according to the quality of the initial attitude. Figures 16-19 plot the EKF estimates of attitude, gyroscope/accelerometer biases and lever arm, respectively, across 50 Monte Carlo runs and Table I lists the final estimate results, as compared with those by the RA-JAPE. We see that the RA-JAPE overwhelmingly outperforms the EKF in all estimates. As shown in Fig. 20, the RA-JAPE's fast convergence property is prominent. The two level angle estimates (roll and pitch) are comparable for both algorithms, while the RA-JAPE is significantly superior to the EKF in the yaw angle estimate. In fact, the yaw estimate by the EKF is badly biased as the mean value stays out of the $\pm 5\sigma$ boundary (see Table I). The RA-JAPE is also remarkably superior to the EKF in terms of accelerometer bias estimate (Fig. 12 vs. Fig. 17) and the lever arm estimate (Fig. 14 vs. Fig. 19). Figures 17-18 indicate the EKF fails to recover any meaningful sensor bias information, which leads to the EKF's inferior performance in attitude estimation. We carry out Monte Carlo simulations to further examine the RA-JAPE's noise sensitivity. Tables II-III respectively summarize the 50 Monte Carlo results for enlarged GPS velocity noise (standard variance 0.2 $m/s$) and for a typical consumer-grade low cost INS, equipped with gyroscopes (drift $10°/h$, noise $36°/h/\sqrt{\text{Hz}}$) and accelerometers (bias $5000\mu g$, noise $80\mu g/\sqrt{\text{Hz}}$). In such significantly high noise cases, although the RA-JAPE still has better mean estimates of the yaw angle and the accelerometer bias, it has lost its overall superiority over the EKF. Admittedly, this observation suggests the RA-JAPE be confined to those accurate applications with high quality INS and GPS measurements, e.g., high-end INS/GNSS reference equipment, georeferencing and gravity measuring.

Both algorithms are implemented using Matlab on a computer with a 2.8G CPU, 4G memory and the Windows 7 operating system. The average running time of 300-seconds-length data is 95 seconds for RA-JAPE and 25 seconds for the EKF. The proposed algorithm requires about four times computation power as much as the EKF does, but is still suitable for online processing.

## VI. Conclusions

INS/GNSS integration is usually carried out in practice by the Kalman-like filtering method, which is prone to attitude initialization failure, especially when the carrier moves freely. This paper has explored a new technique for joint estimation of INS/GNSS attitude and the associated parameters including the GNSS antenna lever arm and inertial sensor biases. The joint estimation problem is posed as a unit quaternion-constrained optimization on attitude, gyroscope/accelerometer biases and lever arm, and a time-recursive Newton-Lagrange algorithm is derived to solve the optimization problem by exploiting its intrinsic property. Numerical results from both the proposed algorithm and a well-tuned EKF have demonstrated that this novel technique has the capability of self-initialization without any prior attitude and sensor noise information, and the proposed algorithm displays significant superiority for those high-end applications with high-quality INS and GPS measurements.

## APPENDIX: INTEGRAL APPROXIMATION IN (23)

Assume the body angular velocity and the specific force to be approximate linear forms as

$$\begin{aligned} \boldsymbol{\omega}_{ib}^{b} &= t\,\mathbf{a}_{\omega} + \mathbf{b}_{\omega} \\ \mathbf{f}^{b} &= t\,\mathbf{a}_{f} + \mathbf{b}_{f} \end{aligned} \tag{53}$$

where $\mathbf{a}_{\omega}$, $\mathbf{b}_{\omega}$, $\mathbf{a}_{f}$ and $\mathbf{b}_{f}$ are the appropriate coefficient vectors. Then we have for the incremental angles

$$\begin{aligned} \Delta\boldsymbol{\theta}_1 &= \int_0^{T/2} \boldsymbol{\omega}_{ib}^{b} dt = \int_0^{T/2} t\,\mathbf{a}_{\omega} + \mathbf{b}_{\omega} dt = \frac{T^2}{8}\mathbf{a}_{\omega} + \frac{T}{2}\mathbf{b}_{\omega} \\ \Delta\boldsymbol{\theta}_1 + \Delta\boldsymbol{\theta}_2 &= \int_0^{T} \boldsymbol{\omega}_{ib}^{b} dt = \int_0^{T} t\,\mathbf{a}_{\omega} + \mathbf{b}_{\omega} dt = \frac{T^2}{2}\mathbf{a}_{\omega} + T\mathbf{b}_{\omega} \end{aligned} \tag{54}$$

from which the coefficient vectors are solved as

$$\begin{aligned} T^2\mathbf{a}_{\omega} &= 4\left(\Delta\boldsymbol{\theta}_2 - \Delta\boldsymbol{\theta}_1\right) \\ T\mathbf{b}_{\omega} &= 3\Delta\boldsymbol{\theta}_1 - \Delta\boldsymbol{\theta}_2 \end{aligned} \tag{55}$$

Similarly for the incremental velocities, we get

$$\begin{aligned} T^2\mathbf{a}_{f} &= 4\left(\Delta\mathbf{v}_2 - \Delta\mathbf{v}_1\right) \\ T\mathbf{b}_{f} &= 3\Delta\mathbf{v}_1 - \Delta\mathbf{v}_2 \end{aligned} \tag{56}$$

Then the second integral on the right side of (23) is approximated by

$$
\begin{aligned}
&\int_{t_k}^{t_{k+1}}\left(\mathbf{I}+\left(\int_{t_k}^{t}\boldsymbol{\omega}_{ib}^{b}d\tau\right)\times\right)dt=\int_{t_k}^{t_{k+1}}\left[\mathbf{I}+\left(\int_{t_k}^{t}\left(\tau-t_k\right)\mathbf{a}_{\omega}+\mathbf{b}_{\omega}d\tau\right)\times\right]dt\\
&=T\,\mathbf{I}+\int_{t_k}^{t_{k+1}}\left(\frac{1}{2}\left(t-t_k\right)^2\mathbf{a}_{\omega}+\left(t-t_k\right)\mathbf{b}_{\omega}\right)\times dt\\
&=T\,\mathbf{I}+T\left(\frac{T^2}{6}\mathbf{a}_{\omega}+\frac{T}{2}\mathbf{b}_{\omega}\right)\times\\
&=T\mathbf{I}+\frac{T}{6}\left(5\Delta\boldsymbol{\theta}_1+\Delta\boldsymbol{\theta}_2\right)\times
\end{aligned}
\tag{57}
$$

As for the first integral of (23), please refer to [10] for its derivation.

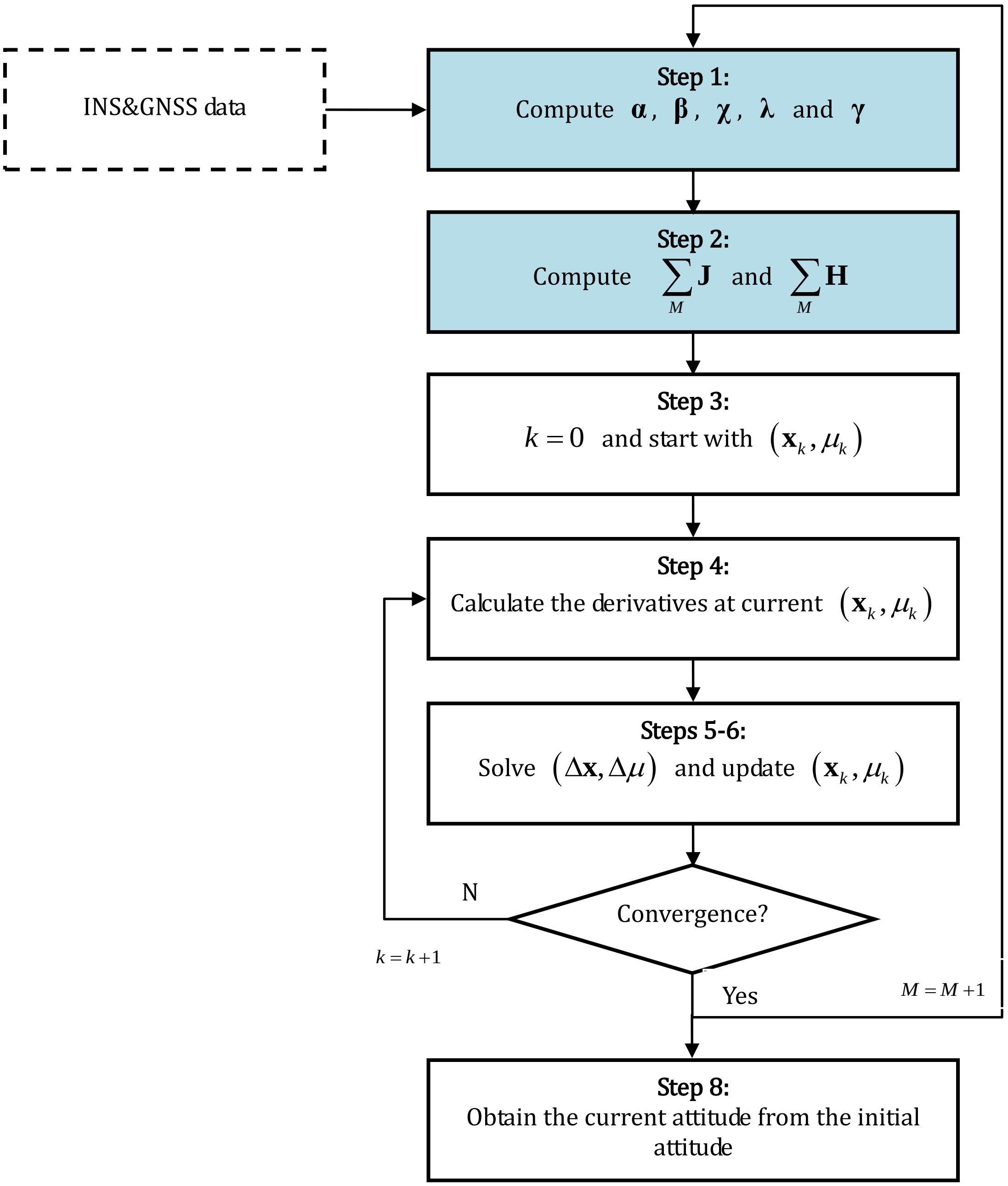

Figure 1. Flowchart of RA-JAPE algorithm (Gray-filled boxes enable incremental computation).

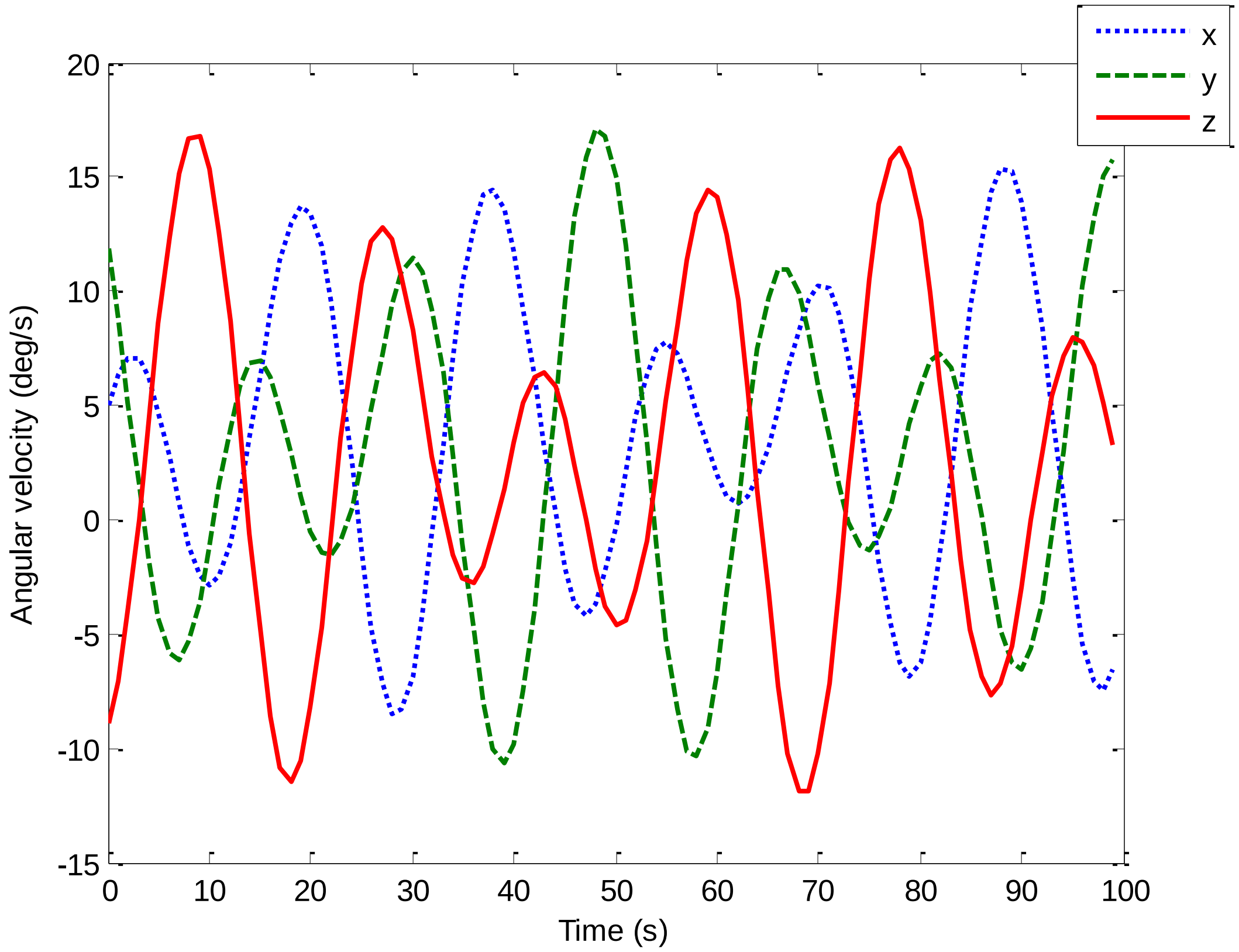

Figure 2. INS angular velocity profile in simulations.

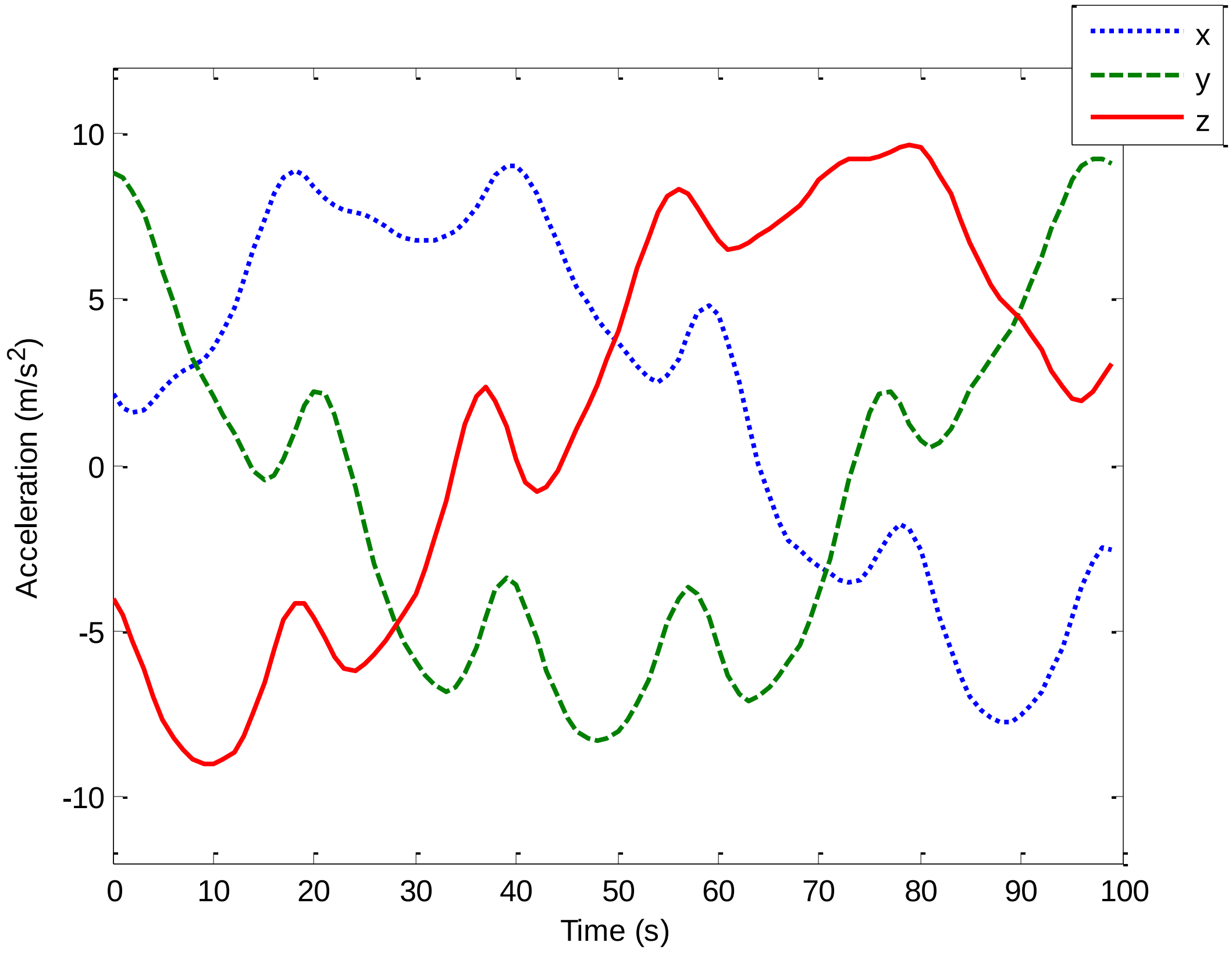

Figure 3. INS acceleration profile in simulations.

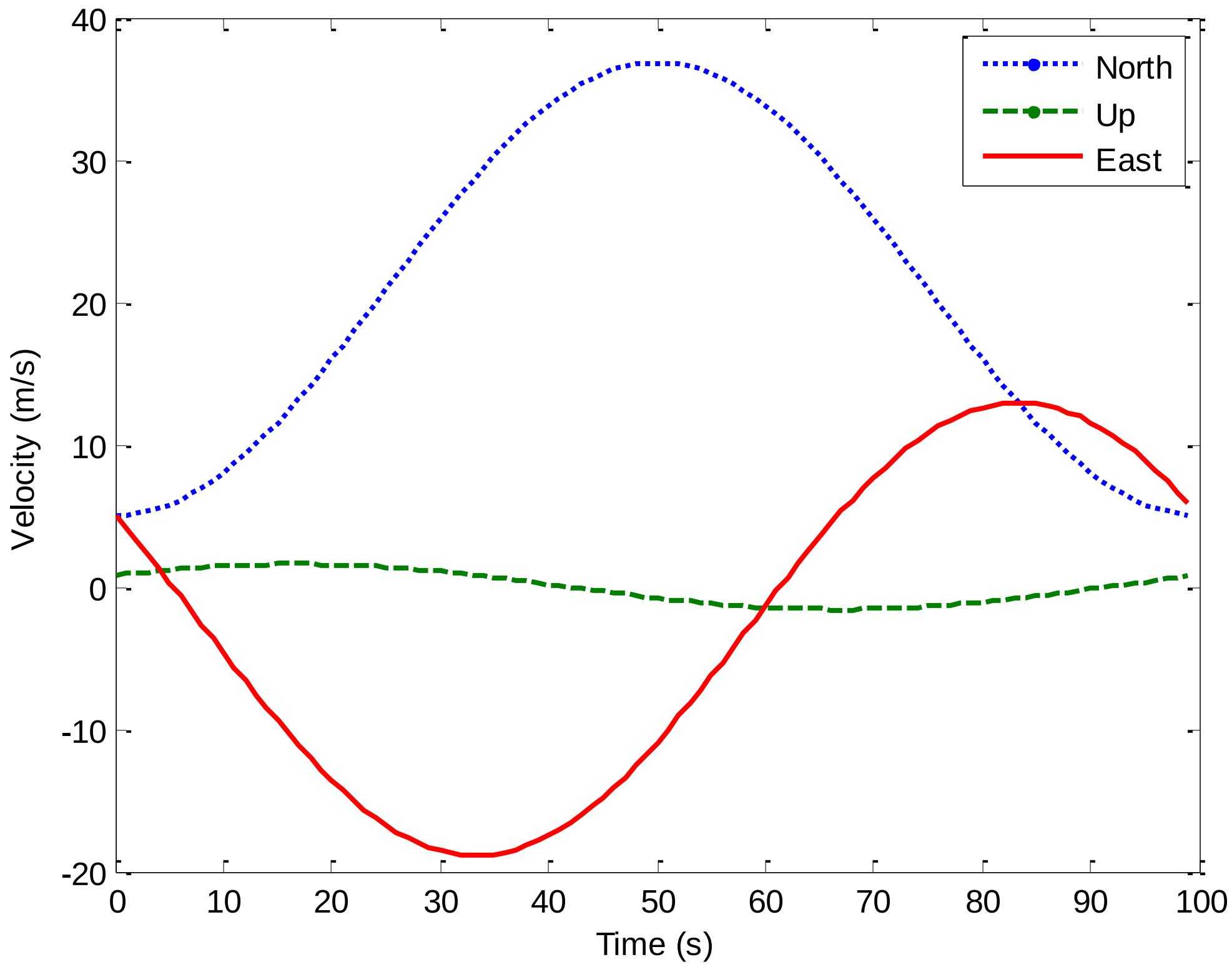

Figure 4. GPS velocity profile in simulations.

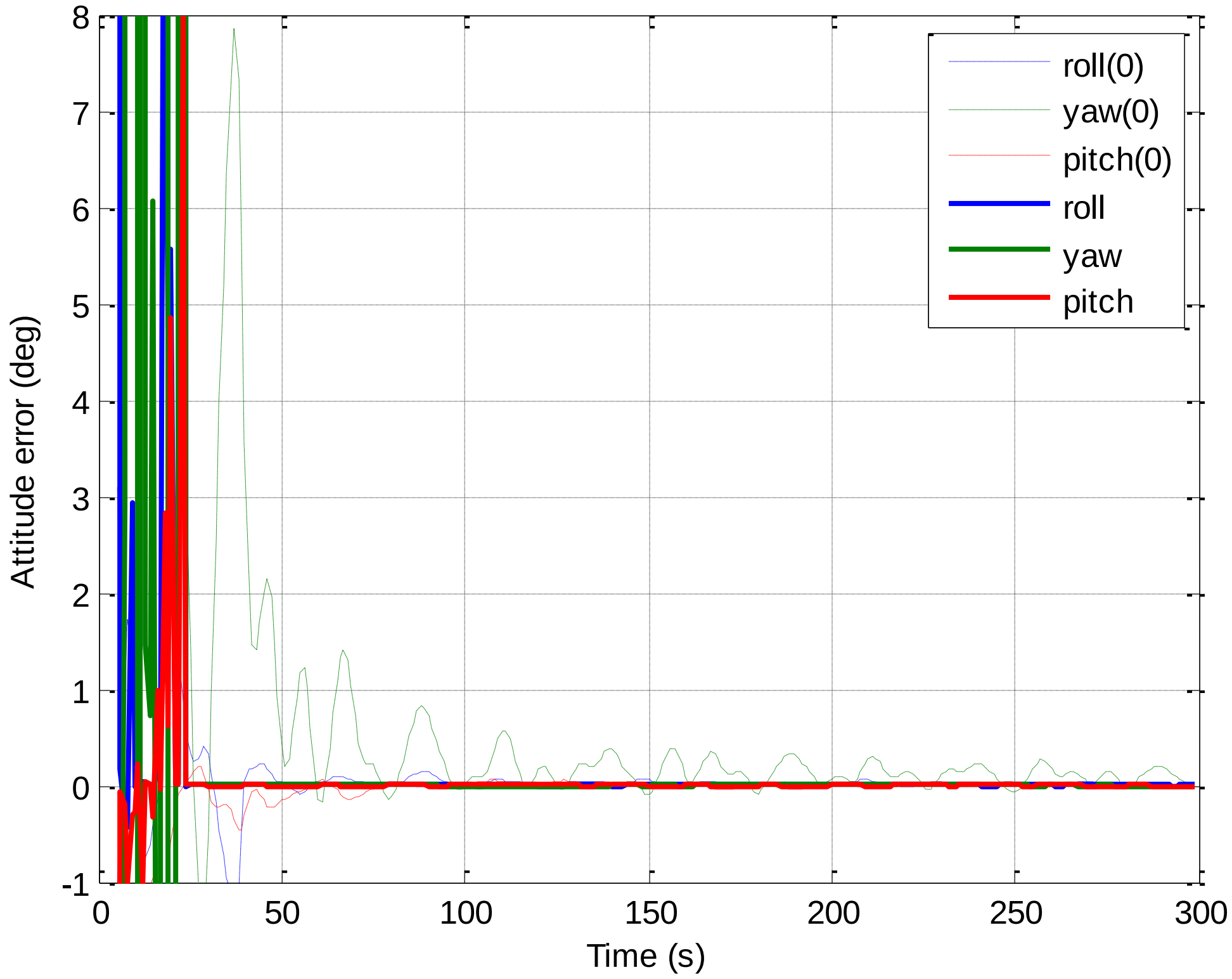

Figure 5. Attitude estimate error (solid lines) in the ideal case, as compared with the initial attitude error (dashed lines).

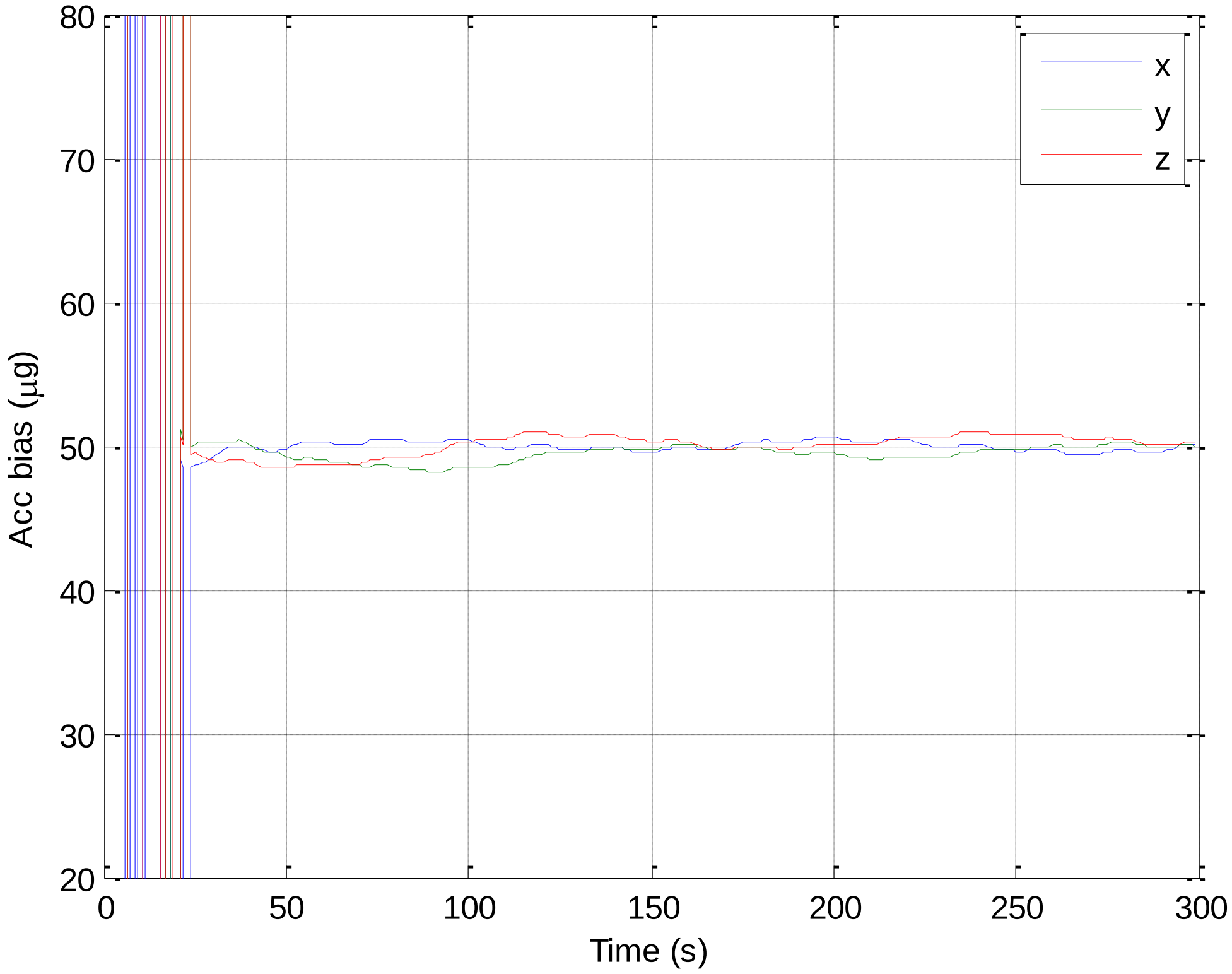

Figure 6. Accelerometer bias estimate in the ideal case.

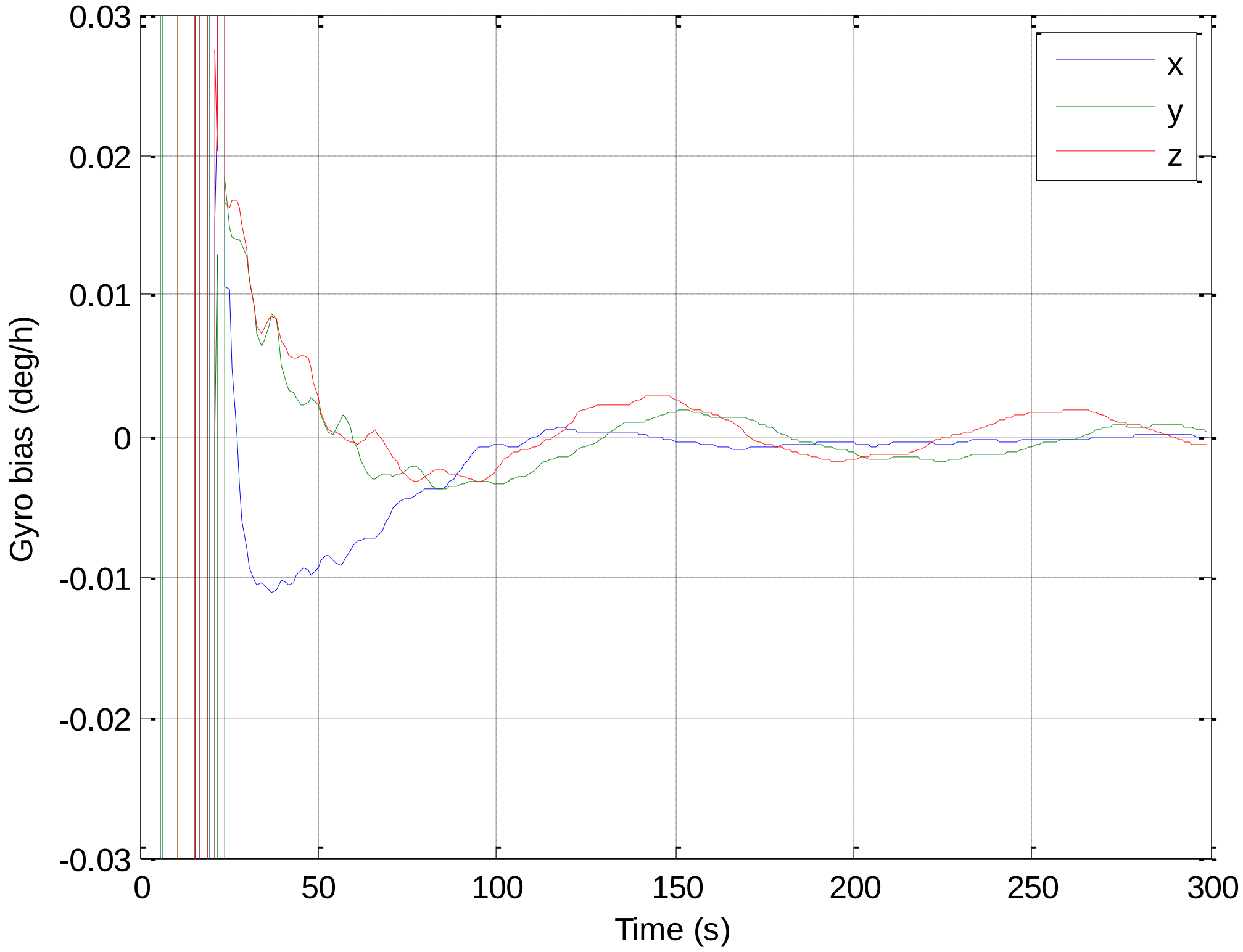

Figure 7. Gyroscope bias estimate in the ideal case.

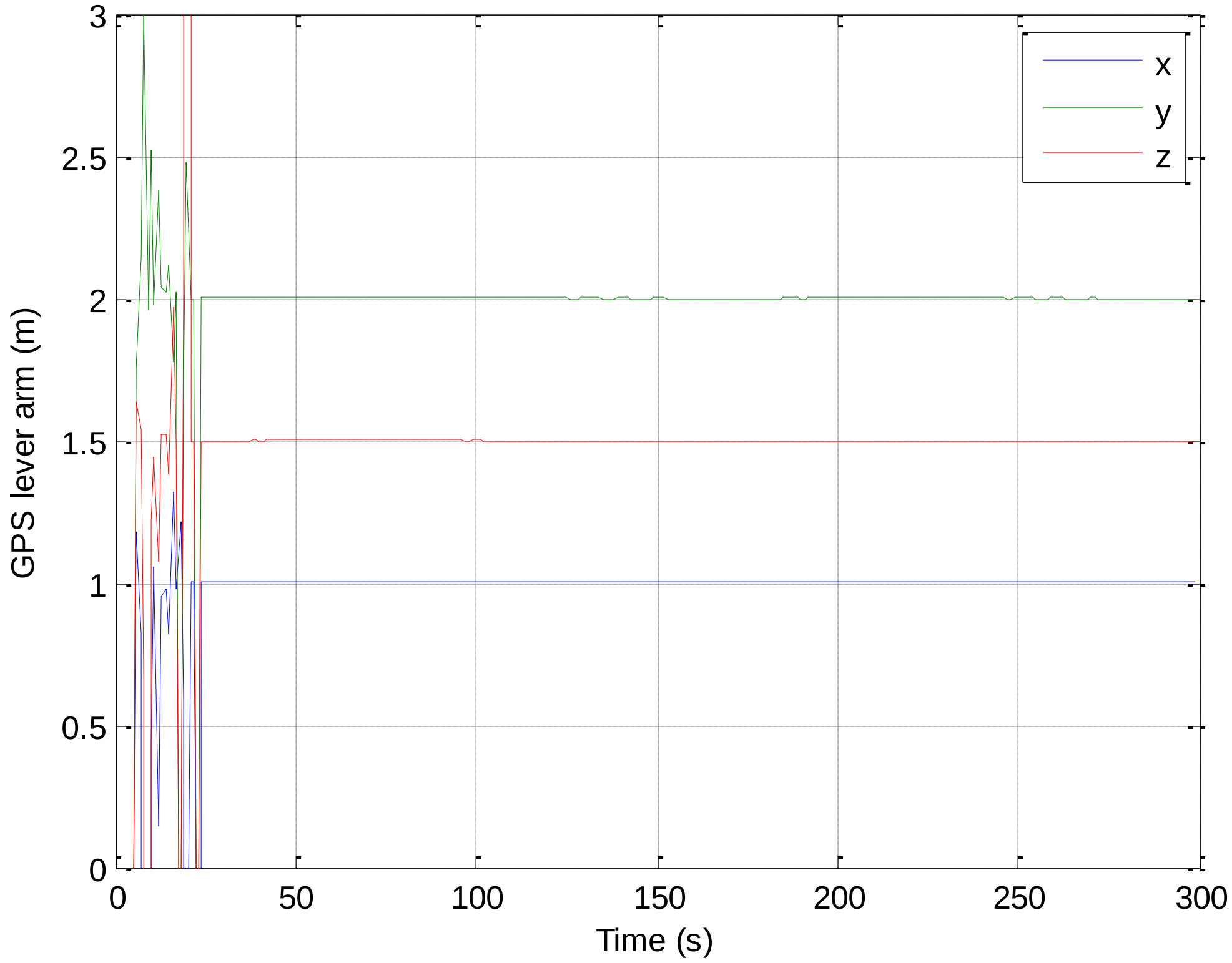

Figure 8. GPS lever arm estimate in the ideal case.

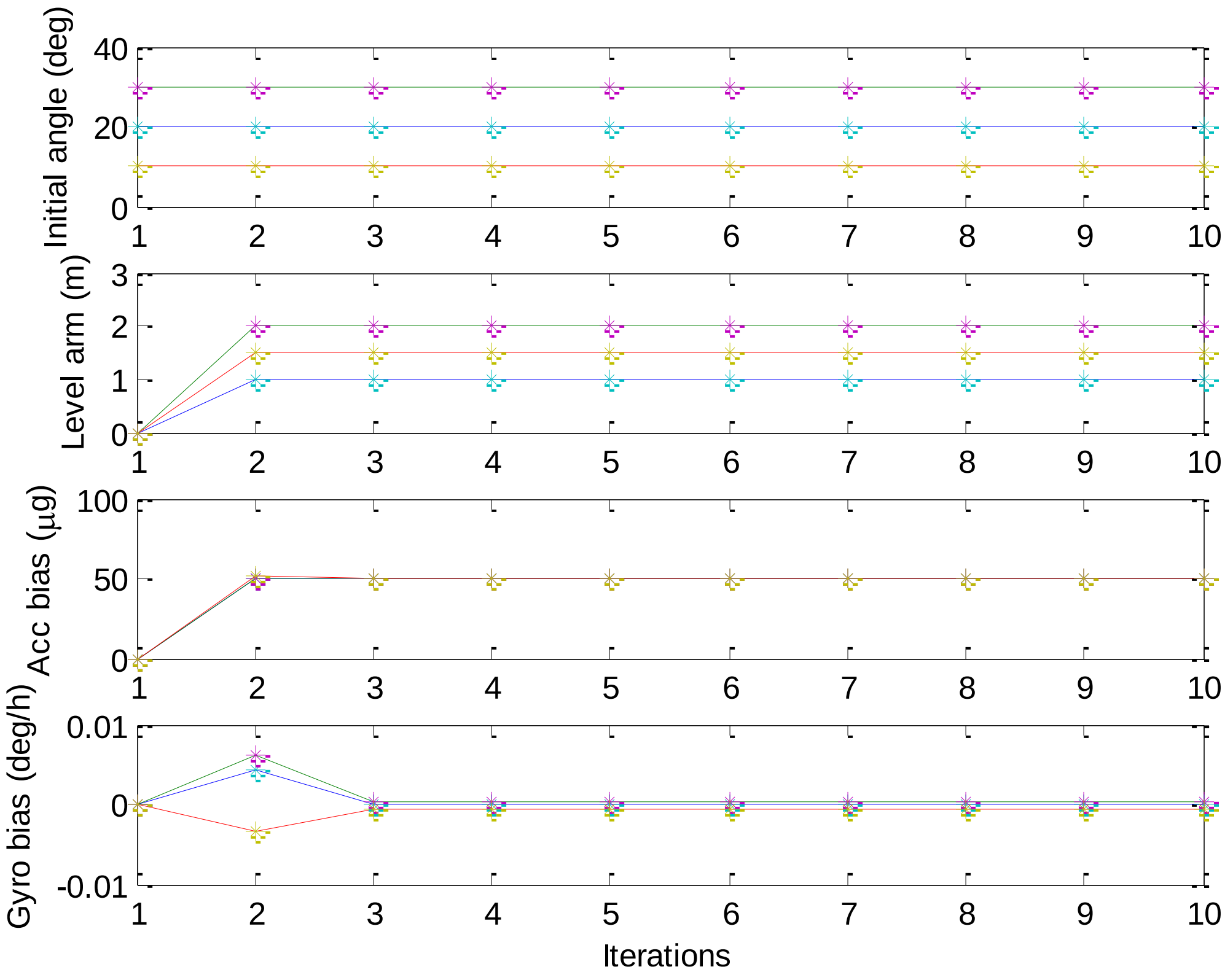

Figure 9. Iterative result of the RA-JAPE algorithm at 300s in the ideal case.

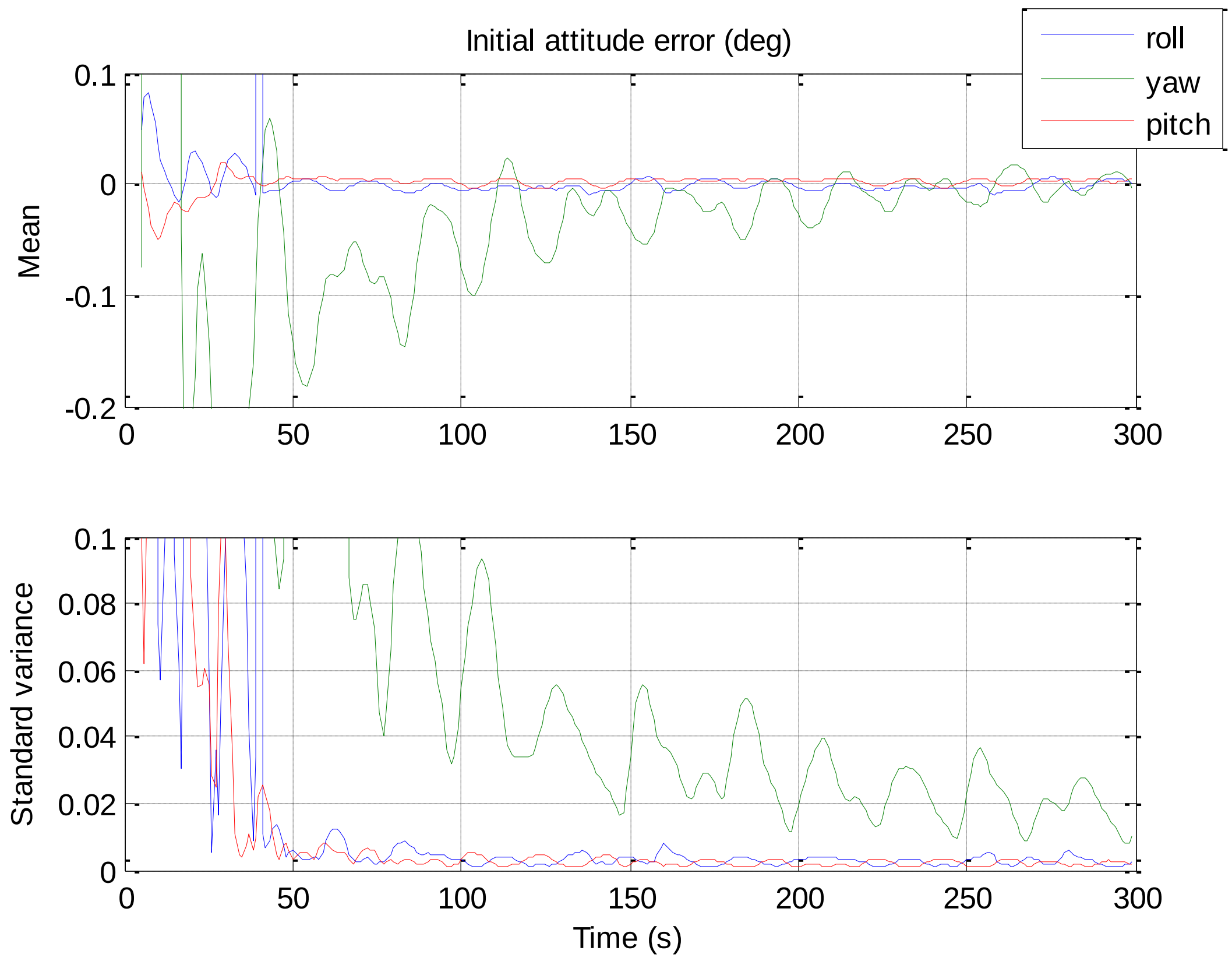

Figure 10. Initial attitude errors across 50 Monte Carlo runs

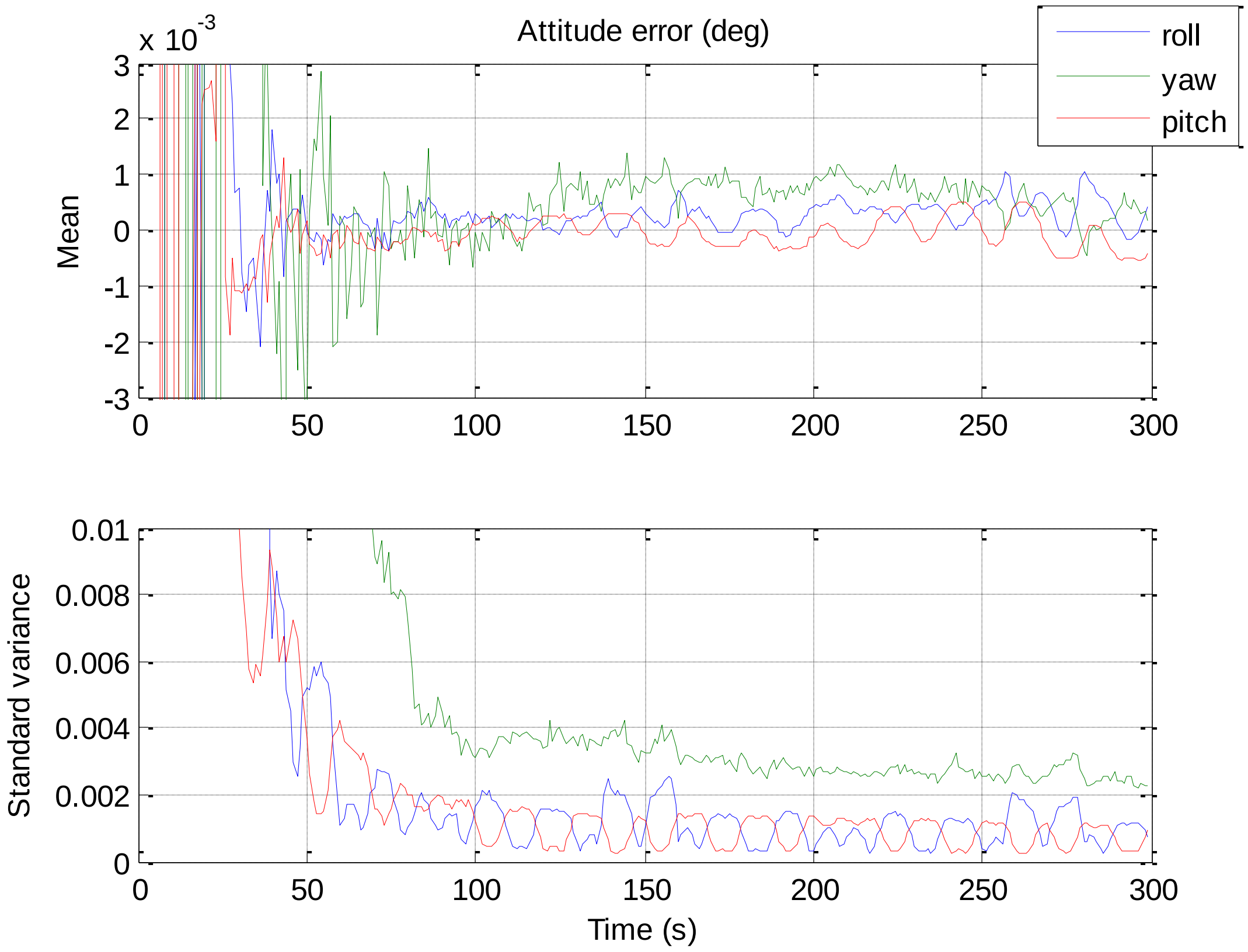

Figure 11. Attitude estimate errors across 50 Monte Carlo runs

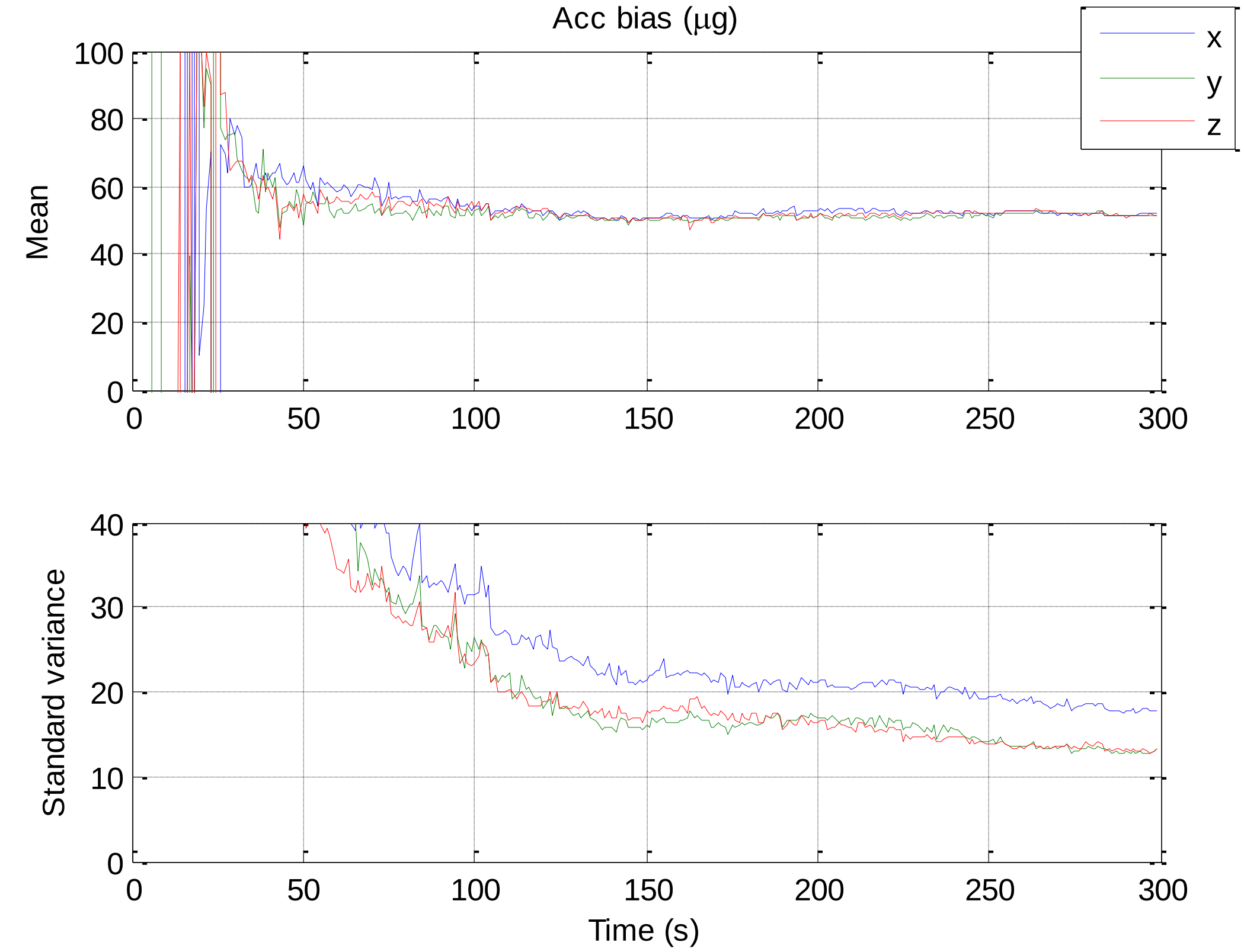

Figure 12. Accelerometer bias estimate across 50 Monte Carlo runs

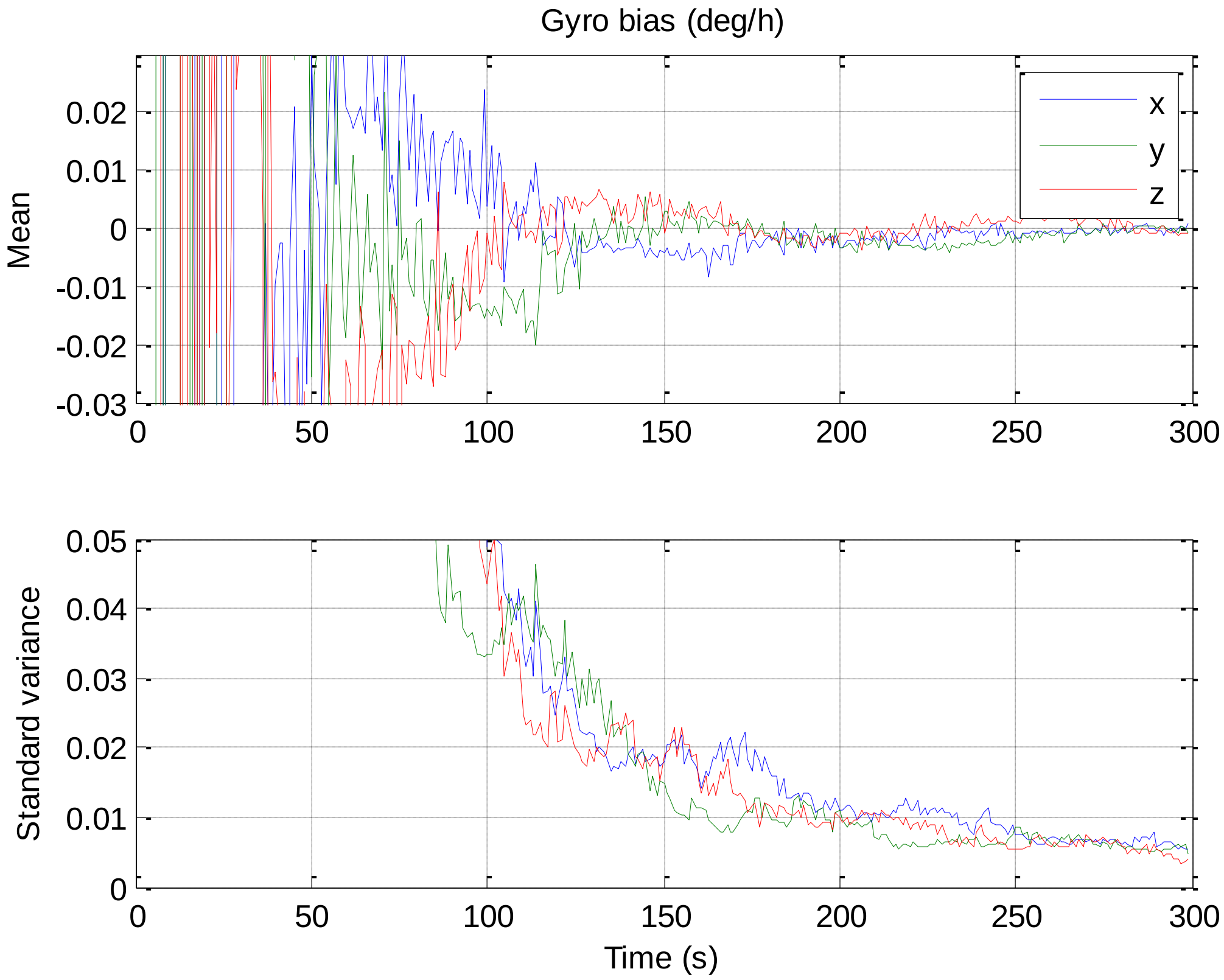

Figure 13. Gyroscope bias estimate across 50 Monte Carlo runs

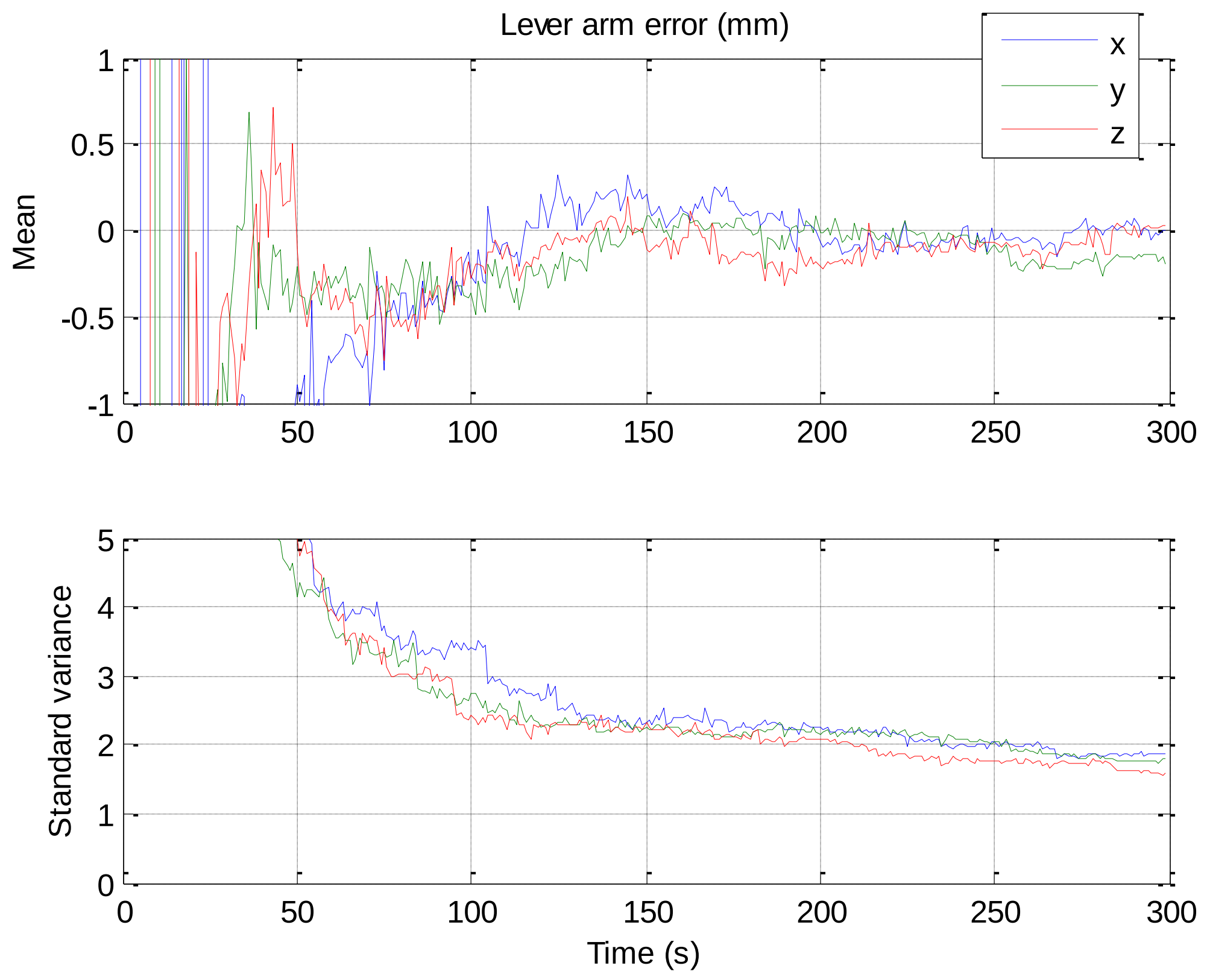

Figure 14. GPS lever arm estimate errors across 50 Monte Carlo runs

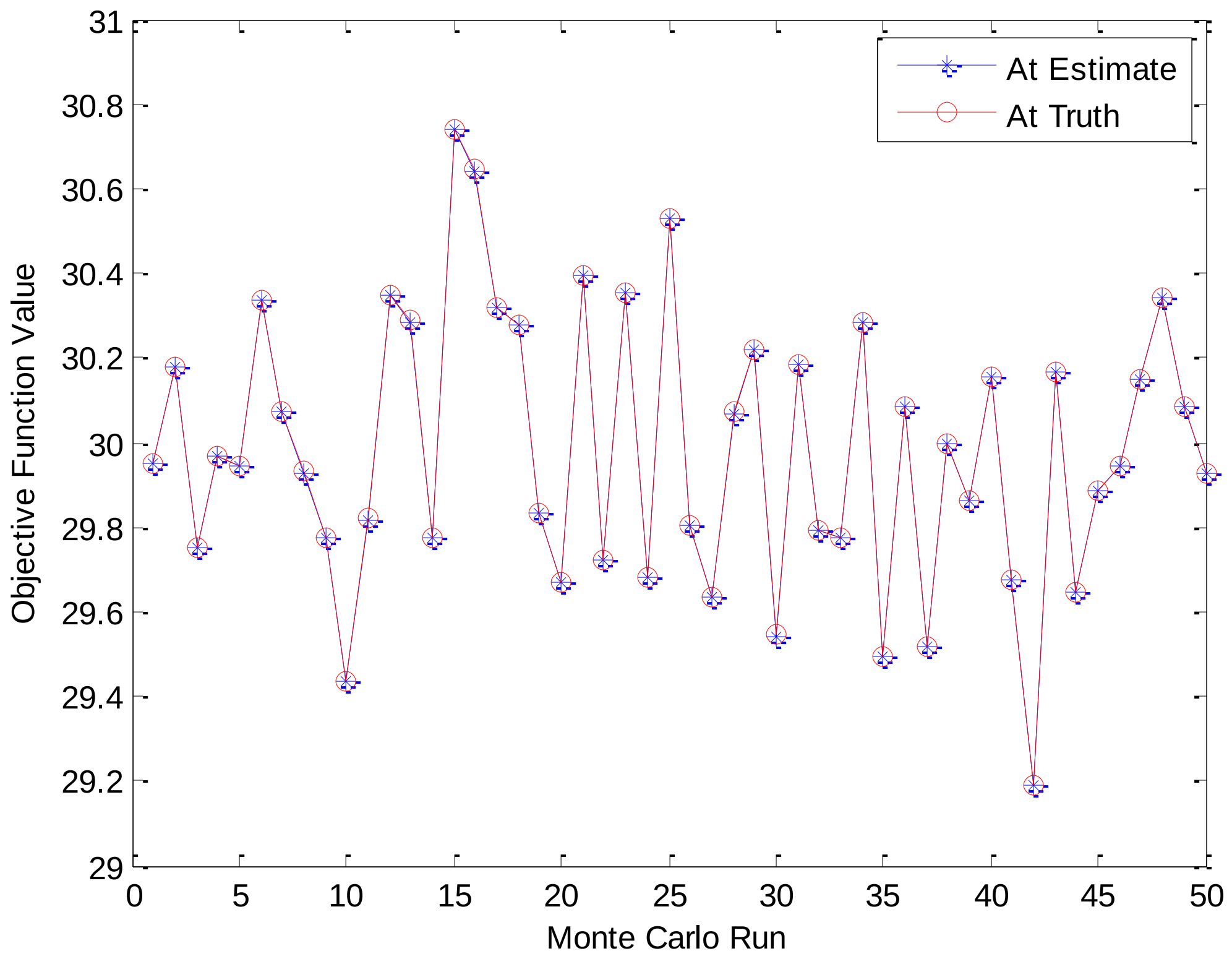

Figure 15. Objective function values at RA-JAPE estimate and the truth across 50 Monte Carlo runs

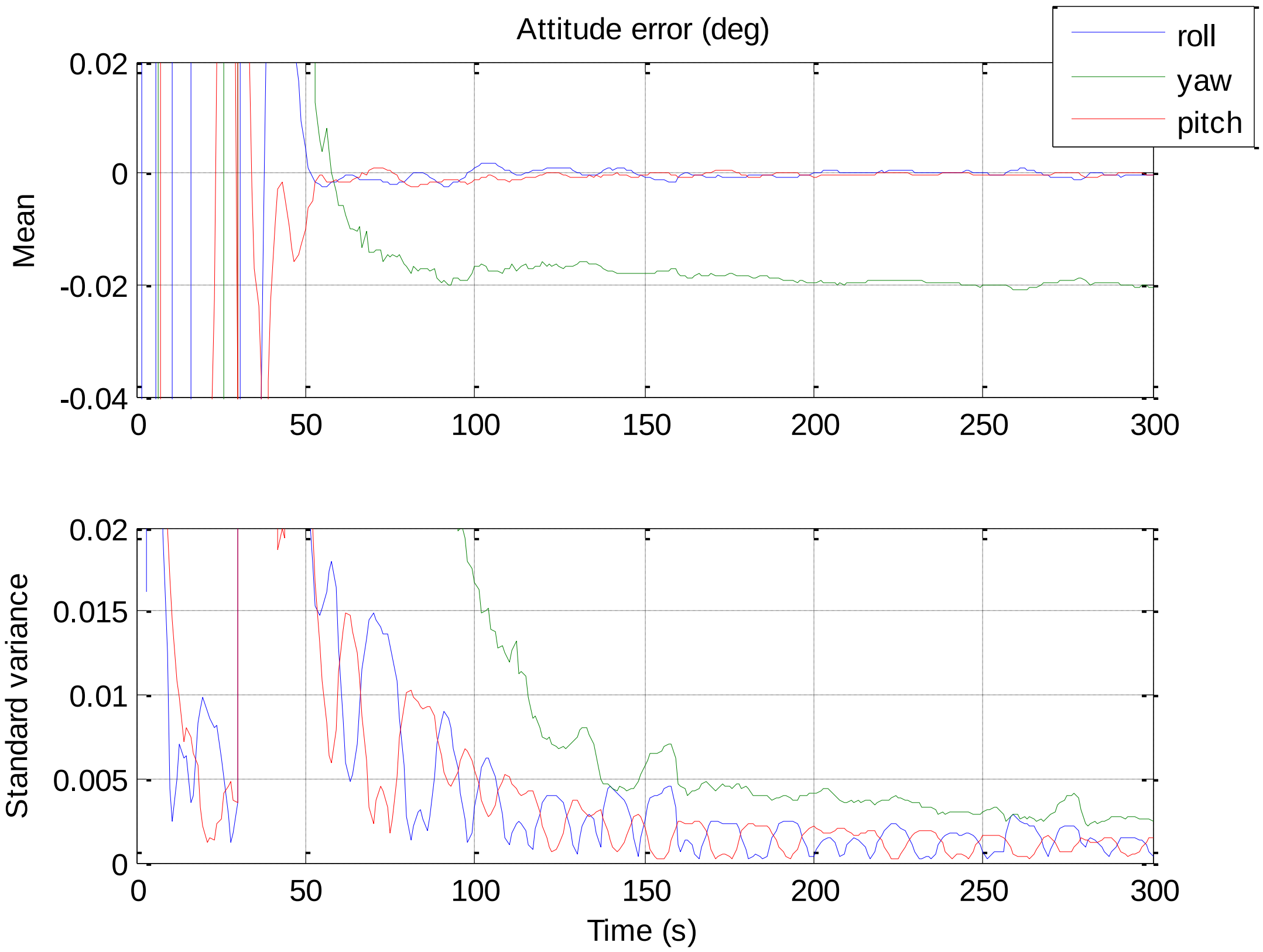

Figure 16. Attitude estimate errors by EKF across 50 Monte Carlo runs

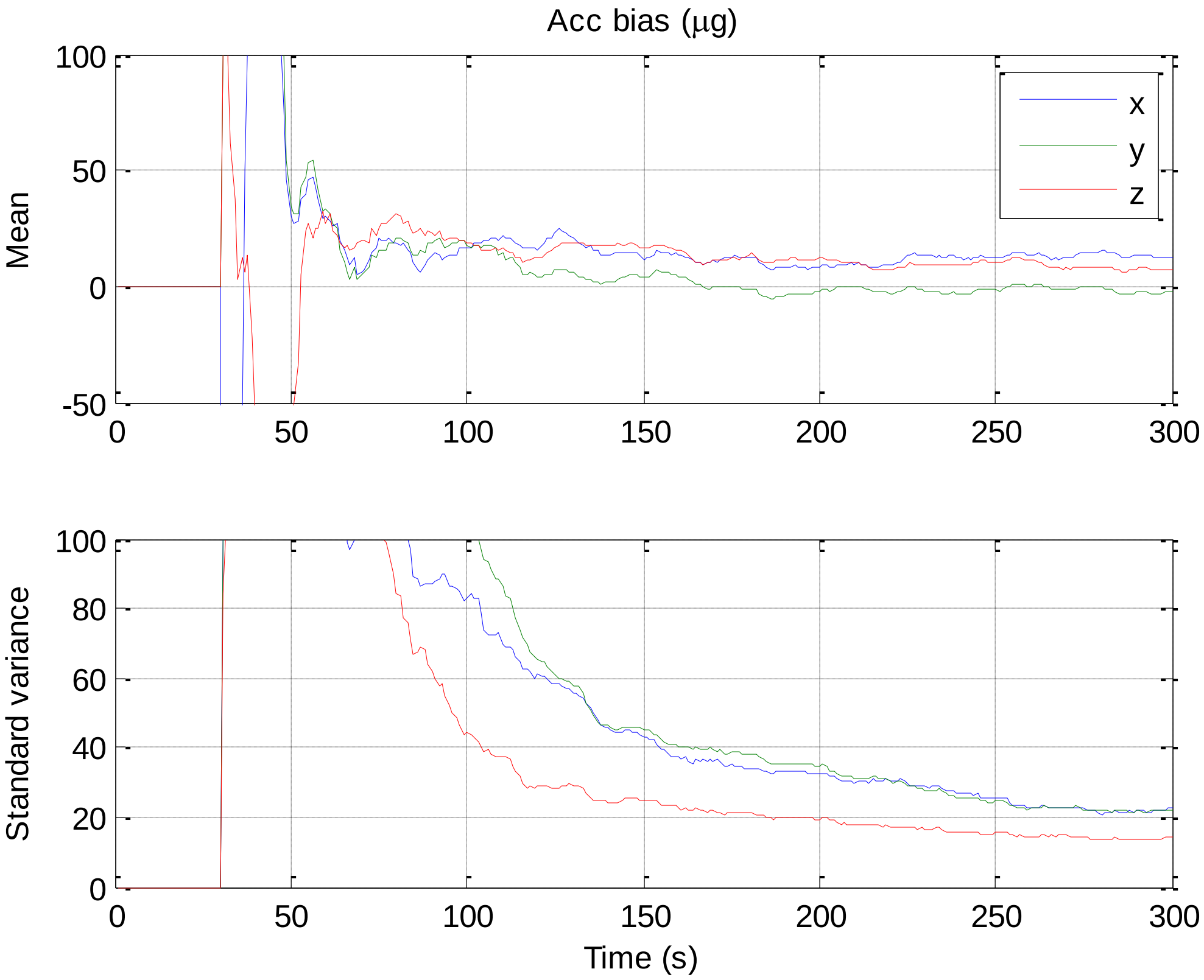

Figure 17. Accelerometer bias estimate by EKF across 50 Monte Carlo runs

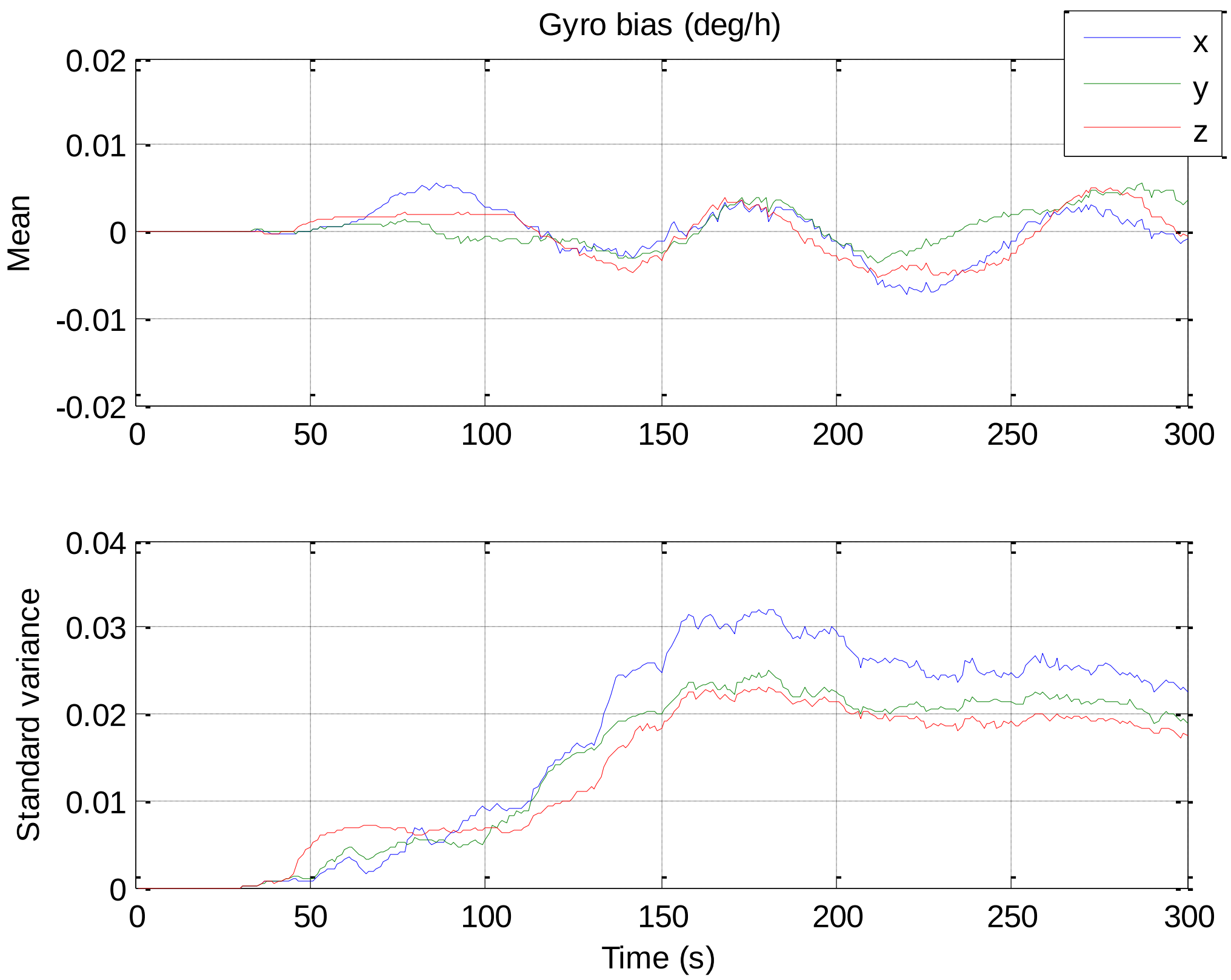

Figure 18. Gyroscope bias estimate by EKF across 50 Monte Carlo runs

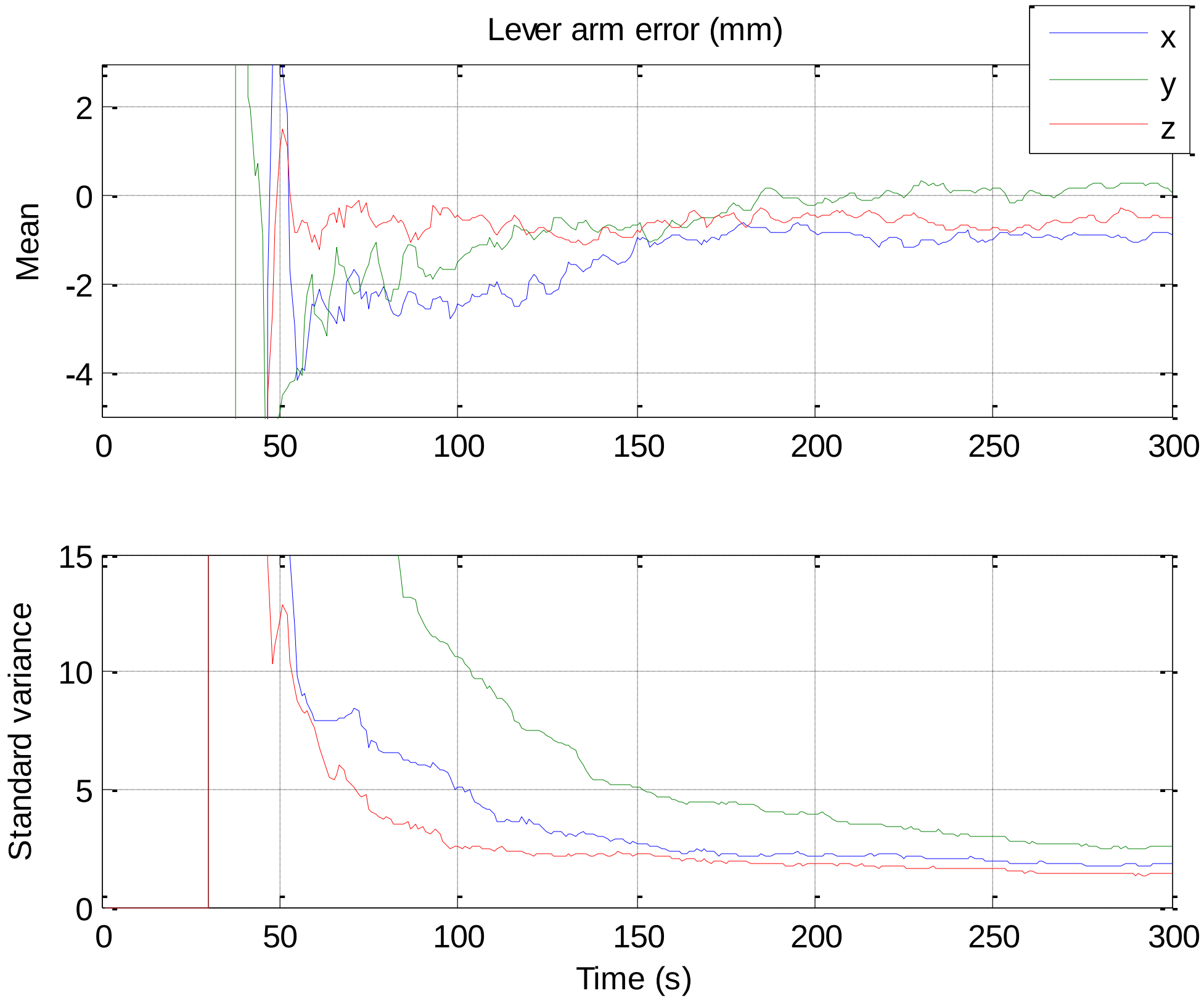

Figure 19. GPS lever arm estimate errors by EKF across 50 Monte Carlo runs

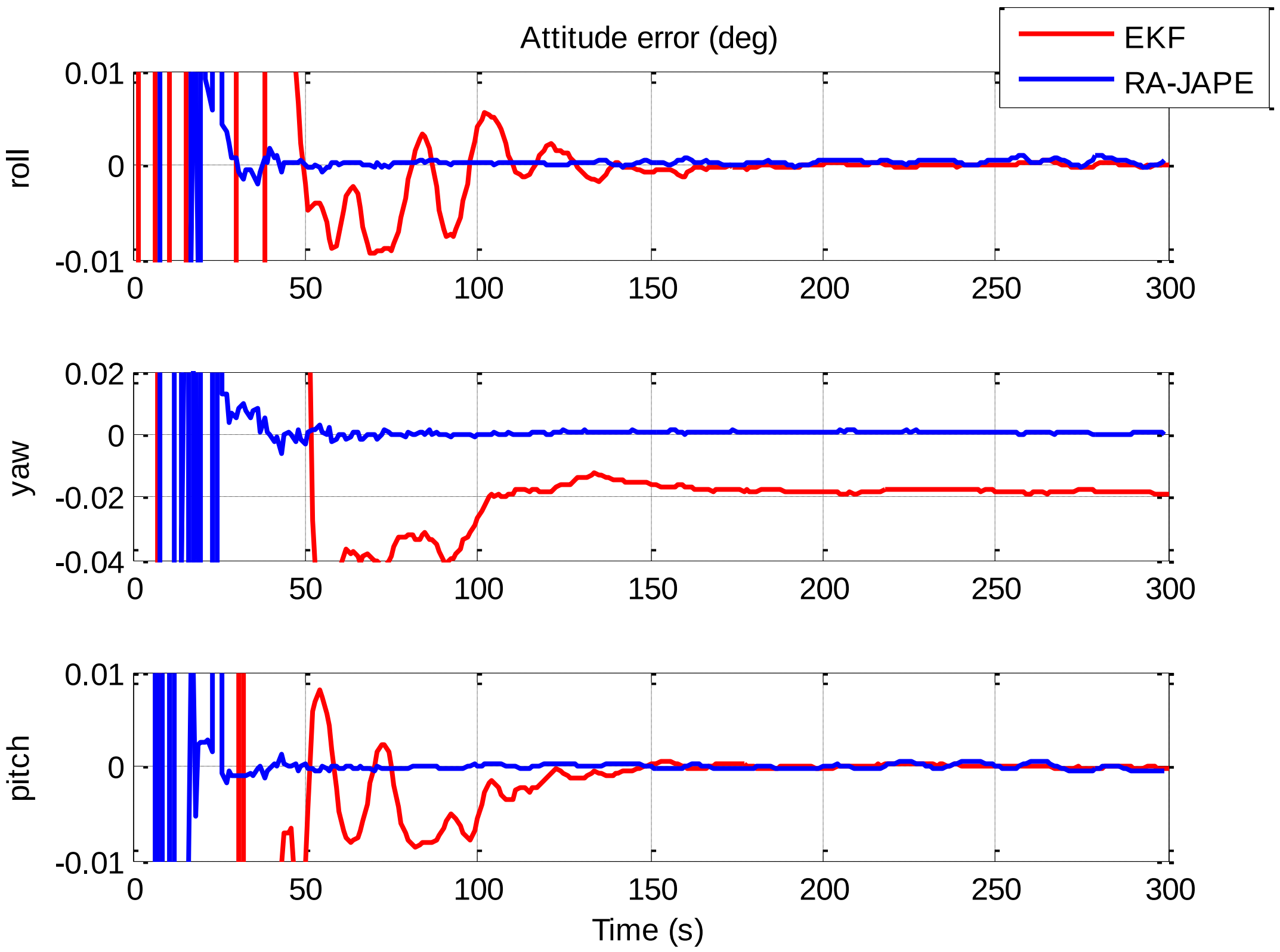

Figure 20. Comparison of attitude mean errors across 50 Monte Carlo runs

Table I. Comparison of Final Estimate Errors (1 σ)

| | **RA-JAPE** | **EKF** |
|---|---|---|
| Attitude (0.001deg) | $[1\pm2 \quad 1\pm3 \quad -0.5\pm2]$ | $[1\pm3 \quad -20\pm4 \quad 0.5\pm2]$ |
| Accelerometer Bias ($\mu g$) | $[2\pm18 \quad 2\pm13 \quad 2\pm13]$ | $[40\pm22 \quad 50\pm14 \quad 40\pm14]$ |
| GPS Lever Arm (mm) | $[0.1\pm1.9 \quad 0.2\pm1.8 \quad 0.1\pm1.6]$ | $[-1\pm2 \quad 0.5\pm3 \quad -0.5\pm1.5]$ |

Table II. Comparison of Final Estimate Errors (1 σ ) for Enlarged GPS Velocity Noise

| | **RA-JAPE** | **EKF** |
|---|---|---|
| Attitude (0.001deg) | $[2\pm10 \quad 2\pm25 \quad -1\pm10]$ | $[1\pm6 \quad -25\pm12 \quad -0.5\pm4]$ |
| Accelerometer Bias ( $\mu g$ ) | $[15\pm160 \quad 5\pm120 \quad 5\pm120]$ | $[35\pm44 \quad 55\pm37 \quad 50\pm30]$ |
| GPS Lever Arm (mm) | $[-5\pm19 \quad 2\pm16 \quad 0.0\pm17]$ | $[-2\pm6 \quad 1\pm7 \quad 1\pm5]$ |

Table III. Comparison of Final Estimate Errors (1 σ ) for a Consumer-grade INS

| | **RA-JAPE** | **EKF** |
|---|---|---|
| Attitude (0.001deg) | $[1000\pm100 \quad -500\pm200 \quad -500\pm100]$ | $[400\pm100 \quad 2000\pm500 \quad 200\pm100]$ |
| Accelerometer Bias ( $\mu g$ ) | $[500\pm1500 \quad 500\pm1500 \quad 1000\pm1800]$ | $[-5100\pm50 \quad -5100\pm25 \quad -5050\pm15]$ |
| GPS Lever Arm (mm) | $[20\pm10 \quad 20\pm10 \quad -50\pm10]$ | $[-20\pm6 \quad 40\pm4 \quad -20\pm4]$ |